\documentclass[12pt,draftcls,onecolumn]{IEEEtran}
\usepackage{amsmath}
\usepackage{cite}
\usepackage{graphicx}
\usepackage{subfigure}
\usepackage{float}
\usepackage{stfloats}
\usepackage{subfloat}
\usepackage{multirow}
\usepackage{array}
\usepackage{lineno}
\usepackage{color}  %
\usepackage{amsmath}
\usepackage{amsfonts}
\usepackage{booktabs}
\usepackage{threeparttable}
\begin{document}
\title{Robust lEarned Shrinkage-Thresholding (REST): Robust unrolling for sparse recovery}

\author{Wei Pu, Chao Zhou, Yonina C. Eldar and Miguel R.D. Rodrigues

\thanks{W. Pu, C. Zhou and M. Rodrigues are with the Department of Electronic and Electrical Engineering, University College London, UK. Y. C. Eldar is with the Weizmann Institute of Science, Rehovot, Israel. This work was supported both by The Alan Turing Institute and Weizmann -- UK Making Connections Programme (Ref. 129589)
}}
	
\maketitle
\begin{abstract}
In this paper, we consider deep neural networks for solving inverse problems that are robust to forward model mis-specifications.
Specifically, we treat sensing problems with model mismatch where one wishes to recover a sparse high-dimensional vector from low-dimensional observations subject to uncertainty in the measurement operator. We then design a new robust deep neural network architecture by applying algorithm unfolding techniques to a robust version of the underlying recovery problem. Our proposed network - named Robust lEarned Shrinkage-Thresholding (REST) - exhibits an additional normalization processing compared to Learned Iterative Shrinkage-Thresholding Algorithm (LISTA), leading to reliable recovery of the signal under sample-wise varying model mismatch. The proposed REST network is shown to outperform state-of-the-art model-based and data-driven algorithms in both compressive sensing and radar imaging problems wherein model mismatch is taken into consideration.
\end{abstract}
\section{Introduction}
\label{sec:intro}
Inverse problems, involving the recovery of a high-dimensional vector of interest from a low-dimensional observation vector, arise in various signal and image processing applications such as compressive sensing\cite{CS1,CS2,CS3}, synthetic aperture radar (SAR)\cite{RI1,RI2}, medical imaging\cite{MI1,MI2,MI3}, and many more.
Three major classes of approaches have been developed over the years to solve such inverse problems. Classical model-based techniques leverage knowledge of the underlying linear model to solve inverse problems via the formulation of optimization problems that include two terms in the objective: (1) a data fidelity term and (2) a data regularization term. The fidelity term encourages the solution to be consistent with the observations whereas the regularization encourages solutions that conform to a certain postulated data prior. For example, variational methods use a regularizer that promotes smoothness of the solutions \cite{Inverse1,Inverse2} whereas sparsity-driven methods use regularizers that promote sparse solutions in some transform domain\cite{CS3,Inverse4}.

More recent data-driven techniques “solve” an inverse problem by using powerful neural networks that learn how to map the model output to the model input based on a number of input-output examples\cite{D1,D2,D3,D4}. These approaches offer various advantages over model based ones including higher performance and lower complexity. \footnote{Data-driven approaches shift the computational burden from the testing phase to the training phase. A neural network -- once optimized -- can deliver an estimate of the target quantity more rapidly than a classical optimization based method.}
However, state-of-the-art data-driven approaches such as deep learning also exhibit various disadvantages: (i) they are often based on heuristic network designs requiring hand-crafted trial-and-error (ii) they typically lack interpretability, including physical meaning; and importantly (iii) they require rich datasets to learn the inverse operation mapping observations to quantities of interest that are not always available.

Finally, in view of the fact that the underlying inverse problem model is often (approximately) known in various settings, there has been increased interest in model-aware data-driven approaches to inverse problems.
A popular model-aware data-driven approach relies on algorithm unfolding or unrolling\cite{Unroll1,Unroll2,Unroll3,Unroll4,Unroll5}. It involves (i) the formulation of an optimization problem allowing to recover the quantities of interest from the observed quantities; (ii) the identification of an iterative algorithm to solve such an inverse problem; (iii) the conversion of such an iterative algorithm onto a network where each algorithm iteration corresponds to a network layer; and (iv) the optimization of the resulting network parameters using back-propagation based on a dataset.  These methods have been applied to a wide range of signal and image processing problems such as compressive sensing\cite{Unroll3}, deblurring\cite{Unroll4}, super-resolution\cite{Unroll6,Unroll7}, and many more, where it has been shown that they can deliver higher performance while offering interpertability in comparison to purely data-driven techniques\cite{r3}.

An increasing challenge arising in inverse problems affecting the performance of these various approaches relates to the presence of mismatch between the actual model and the postulated one due to erroneous modelling of the forward model\cite{PCS1,PCS2} or other anomalies. For example, in applications such as SAR imaging such a mismatch may arise from the motion error of the aircraft due to air disturbance. Likewise, in applications such as MRI imaging mismatch may arise from unexpected movement or jitter of the target object.

In scenarios where the mismatch between the true model and the postulated one is fixed, it has been shown that model-based approaches to inverse problems can suffer considerable degradation\cite{PCS2}; in contrast, data-driven techniques (including model aware data driven ones such as learnt iterative soft threshold algorithm (LISTA)\cite{Unroll1}) are less affected because the inverse mapping corresponding to the actual model can still be learnt given enough training data.
However, when the mismatch between the true model and the postulated one is not fixed but varies potentially across measurements over time -- such as in the aforementioned SAR and MRI imaging settings -- the performance of data-driven techniques, including model-aware ones, can be severely compromised. In fact, it has been shown that, in the presence of mismatch, the performance of model-based approaches to inverse problems may be significantly better than existing deep learning based methods\cite{r3}.

Therefore, a number of approaches have recently emerged to successfully tackle model mismatch in inverse problems. For example, \cite{PCS3} proposed a total-least squares recovery algorithm allowing the reconstruction of a high-dimensional vector from low-dimensional noisy measurements in the presence of mismatch in the forward measurement operator. In \cite{PCS4}, the authors proposed a matrix-uncertainty generalized approximate message passing algorithm in a Bayesian framework with informative priors allowing the reconstruction of high-dimensional vectors with random mismatch.
There also exists various approaches to deal with model mismatch in different practical applications. The authors in \cite{r1} formulated a dynamic magnetic resonance imaging problem as a blind compressive sensing problem, wherein the forward-model is assumed to be integrated with unknown mismatch.
In SAR applications, an alternating minimization strategy was proposed in \cite{r2} to simultaneously estimate the model mismatch and the sparse coefficients from the SAR data.

However, to the best of our knowledge, there has been less progress in the development of data-driven approaches, including deep learning ones, to inverse problems that are robust to model mismatch.
There are some works relating to the design of deep neural networks robust to adversarial attacks\cite{AA1}, which are small imperceptible perturbations to the input data that seriously degrade the performance of an otherwise high-performing data-driven approaches. These include techniques based on adversarial training\cite{AA2}, data pre-processing\cite{AA3} and robust neural network design\cite{AA4,AA5}. However, robustness against model mismatches has not been addressed.

In this paper, we propose a new approach to design robust neural network architectures, leveraging unfolding techniques. This allows to solve the linear inverse problem subject to uncertainty in the measurement operator.
In particular, by building upon a total least squares formulation of a inverse problem where one wishes to recover a high-dimensional vector from low-dimensional noisy measurements subject to model mismatch, we make the following contributions:

\begin{itemize}
  \item We first develop a (robust) iterative soft threshold algorithm (ISTA) based solver to the underlying total least squares problem using conventional proximal gradient.
  \item We then map this robust ISTA solver onto a Robust lEarned Shrinkage Thresholding (REST) network using algorithm unfolding techniques whereby each robust ISTA iteration corresponds to a REST network layer.
  \item We demonstrate via experiments that REST can outperform various model-based and data-driven algorithms in compressive sensing and radar imaging problems.
\end{itemize}


The remainder of the paper is organized as follows: Section II formulates the linear inverse problem in the presence of mismatch in measurement operator. Section III describes the design of our proposed REST network. Sections IV and V present experimental results in compressive sensing and radar imaging tasks, respectively. Finally, in section VI, we draw conclusions.
The Appendix A presents experimental results in compressive sensing task to compare the performance of the proposed REST network with shared and unshared parameters in each layer.
The Appendix B presents some alternative strategies to unfold REST network along with their performance in a simple compressive sensing task.
The Appendix C presents experimental results in compressive sensing task to support our intuitions on the comparison between REST and LISTA.

\section{Problem formulation}

Consider a conventional linear inverse problem given by:

\begin{align}
\label{e1}
y = A x +e,
\end{align}
where $y \in {\mathbb R}^{M \times 1}$ is an observation vector, $x \in {\mathbb R}^{N \times 1}$ is the vector of interest, $e \in {\mathbb R}^{M \times 1}$ is measurement noise, and $A \in {\mathbb R}^{M \times N}$ models the inverse problem forward operator where $N > M$. The goal is to reconstruct $x$ from the observation $y$ given a measurement matrix $A$. These problems arise in various domains in signal and image processing, such as medical imaging systems\cite{IVP1}, high-speed videos\cite{IVP2}, single-pixel cameras\cite{IVP3}, and radar imaging\cite{IVP4}.

It is not generally possible to recover $x$ from $y$ when $N > M$ unless one makes additional assumptions about the structure of the vector of interest. In particular, by postulating that $x$ is sparse, a popular approach to recover $x$ from $y$ involves using the least-absolute shrinkage and selection operator (Lasso)\cite{LASSO}, which is a solution to
\begin{align}
\label{e2}
\min_x & \quad \| y - Ax\|_2 + \lambda \| x \|_1,
\end{align}
where $\| \cdot \|_2$ is the $l_2$ norm and $\| \cdot \|_1$ denotes the $l_1$-norm. This optimization problem can be solved using the well-known iterative soft thresholding algorithm (ISTA)\cite{ISTA} or ADMM\cite{ADMM}.

Alternatively, one can adopt algorithm unfolding or unrolling techniques\cite{Unroll4} to map such solvers onto a neural network architecture, whose parameters can then be further tuned using gradient descent or some variant based on the availability of a series of examples $\{(x_i,y_i\}_{i=1}^{n}$. Networks derived from ISTA or ADMM known as LISTA\cite{Unroll1} and ADMM-Net\cite{Unroll3} respectively, have been shown to perform much better than purely learnt networks (e.g.\cite{D4}) or ISTA\cite{ISTA} and ADMM\cite{ADMM}.

Consider now a more challenging scenario where the observation vector $y \in {\mathbb R}^{M \times 1}$ is related to $x \in {\mathbb R}^{N \times 1}$ as:
\begin{align}
\label{e3}
y = (A+E) x +e.
\end{align}
The model in (\ref{e1}) differs from the model in (\ref{e3}) in that the forward operator $A$ - which is assumed to be known - is now contaminated by an error matrix $E$, which is unknown. Therefore, one now needs to recover the vector of interest in the presence of model mismatch $E$.

To recover $x$ from $y$ in this case, one may consider a robust version of LASSO, or an $l_1$ regularized version of total least squares\cite{PCS3} as follows:
\begin{align}
\label{e4}
\min_{x,E} & \quad \| [E \quad e ]\|_2 + \lambda \| x \|_1, \nonumber\\
{\rm s.t.}& \quad  y =  (A+E)x + e .
\end{align}
Our goal is to build upon this formulation in order to design a data-driven approach using algorithm unfolding or unrolling techniques to recover $x$ reliably from $y$ in the presence of model mismatch. Model mismatch can be divided into two categories according to whether the mismatch is varying or fixed for different samples:
\begin{enumerate}
  \item {\textbf {sample-wise fixed mismatch:}} In this scenario, the model mismatch is fixed for different data samples. One can then reliably recover $x$ from $y$ using data driven approaches such as LISTA. The reason is that the inverse mapping corresponding to the actual model can be learnt given enough training data as varying samples share the same model mismatch.
  \item {\textbf {sample-wise varying mismatch:}} In this scenario, the model mismatch is changing for different data samples, which is much more common in practical applications. It is very hard for existing data-driven approaches to deal with sample-wise varying model mismatch. There also appears to be less work on how to design data-driven approaches -- especially deep learning ones -- with sample-wise varying model mismatch.
\end{enumerate}
Our goal is to design a data-driven approach using algorithm unfolding technique based on the formulation in (\ref{e4}) for sample-wise varying mismatch.

\section{Robust Learned Shrinkage-Thresholding (REST) Network}

\subsection{Robust ISTA}

As shown in \cite{PCS3}, the problem (\ref{e4}) can be solved in two different ways:
\begin{enumerate}

\item The first approach involves simultaneous recovery of  $x$ and $E$ given $y$. This is possible by converting (\ref{e4}) into
    \begin{align}
    \label{e5}
    \min_{x,E} & \quad \| y- (A+E)x\|^2_2 +\| E \|_F^2 + \lambda \| x \|_1.
    \end{align}
\item By admitting the closed-form solution of $E$, the second approach converts the original total least squares optimization problem in (\ref{e4}) into the following optimization problem only related to $x$:
    \begin{align}
    \label{e7}
    \min_{x} \frac{\| y-Ax \|^2_2}{1+\| x \|^2_2}  + \lambda \| x \|_1.
    \end{align}
\end{enumerate}


In \cite{PCS3}, the optimization problem in (\ref{e7}) is solved by a two iteration loop method, which comprises an outer iteration loop based on the bisection method\cite{PCS3-1}, and an inner iteration loop that relies on a variant of branch-and-bound (BB) method\cite{PCS3-2}.
Here, we propose a proximal gradient method in order to design an iterative algorithm to recover $x$ from $y$, which will be utilized to unfold into a network.
This proximal gradient method here can be seen as a robust version of the ISTA algorithm, because both ISTA and robust ISTA solve a $l_1$ regularized version of a least square problem using proximal gradient methods. However, robust ISTA deals with the optimization problem appearing in (\ref{e7}), where model mismatch is present, whereas ISTA deals with the optimization problem in (\ref{e2}).

Taking the gradient on the first term in (\ref{e7}) and executing a proximal step on the second term results in
\begin{align}
\label{e8}
x^{k+1}= {\cal S}_{\mu \lambda} \left\{   x^k - \mu g^k \right \},
\end{align}
where $\mu > 0$ is a step size, $x^{k}$ represents the estimation of $x$ in the $k$-th iteration, and $g^k$ denotes the gradient of the first term evaluated at $x^k$:
\begin{align}
\label{e9}
g^{k} = \frac{2\left[ (1+ \|{  x^k} \|_2^2) A^T(Ax^k-y)-\| y - Ax^k \|_2^2x^k \right]} {(1+ \|{  x^k} \|_2^2)^2}.
\end{align}
The operator $S_{\mu \lambda} (\cdot)$ is the soft thresholding operator, and is applied element-wise on its vector argument as
\begin{align}
\label{e9.1}
{\cal S}_{\mu \lambda}\{ x \} = {\rm sign}(x)\cdot \max (|x| -\mu \lambda , 0).
\end{align}

We next adopt unfolding techniques in order to map the robust recovery algorithm in (\ref{e8}) onto a robust neural network that can be used to recover high-dimensional sparse vectors from low-dimensional noisy measurements in the presence of model mismatch.

\subsection{Robust Learned Shrinkage-Thresholding Network}

We first use (\ref{e9}) in order to re-write (\ref{e8}) as follows:
\begin{align}
\label{e10}
&x^{k+1} = \nonumber\\
&{\cal S}_{\mu \lambda} \left ( x^{k} + \frac{2\mu\| y - Ax^k \|_2^2} {(1+ \|{  x^k} \|_2^2)^2} x^{k} - \frac{2\mu A^TA} {1+ \|{  x^k} \|_2^2 } x^{k}  +  \frac{2\mu A^T} {1+ \|{  x^k} \|_2^2  } y\right).
\end{align}
Replacing $A$ in the first term by $A_1$, $A^TA$ in the second term by $A_2$, and $A^T$ in the third term by $A_3$, (\ref{e10}) becomes
\begin{align}
\label{e11}
&x^{k+1} = \nonumber\\
&{\cal S}_{\mu \lambda} \left (  x^{k} + \frac{2\mu\| y - A_1x^k \|_2^2} {(1+ \|{  x^k} \|_2^2)^2} x^{k} - \frac{2\mu A_2} {1+ \|{  x^k} \|_2^2 } x^{k}  +  \frac{2\mu A_3} {1+ \|{  x^k} \|_2^2  } y\right) .
\end{align}

We now map the iterative algorithm in (\ref{e11}) onto a neural network architecture using unfolding techniques.
In particular, we firstly map each iteration of the algorithm onto a layer of the neural network. Such a layer produces an output $x^{k+1}$ -- which is input to the next layer -- based on an input $x^k$ that undergoes a series of transformation such as linear transformations, linear scalings, and soft-thresholding operations.
We then stack various such layers in order to produce an overall neural network. Such a network consisting of $K$ layers corresponds to $K$ iterations of the original robust ISTA algorithm, it accepts as input an initialization $x^0$, and it delivers as output an approximate solution $x^K$.
Finally, we use different learnable parameters per network layer corresponding to the original parameters. Concretely, we choose learnable parameters $\lambda, \mu, A_1, A_2$ and $A_3$.
Note that each layer in the proposed network share the same learnable parameters (i.e., $\lambda, \mu, A_1, A_2$ and $A_3$) are set to be the same in each layer) to give the network more restrictions to promote a better performance. We compare the performance of shared and unshared learnable parameters in each layer of REST in Appendix A.

\begin{figure}[h]
\centering
\includegraphics[width=0.65\textwidth]{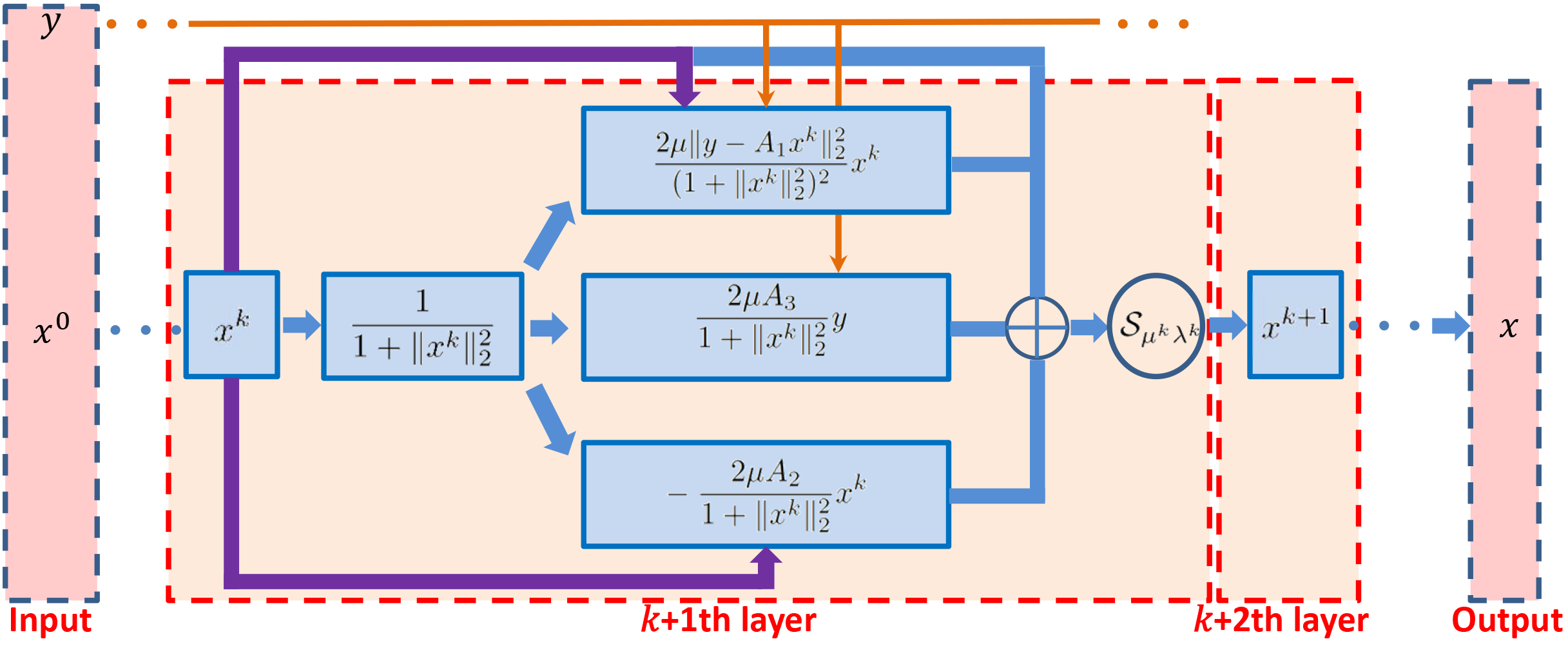}
\caption{REST Neural Network Architecture.}\label{fig1}
\end{figure}

The architecture of our network is depicted in Fig. \ref{fig1}.
We note that depending on how we define the learnable parameters associated with (\ref{e10}) and (\ref{e11}) we may have slightly different neural network architectures. We elaborate on these possibilities in Appendix B.
\subsubsection{Learning Algorithm}

To optimize the learnable parameters we rely on a dataset containing a series of pairs ($x_l,y_l$), $l=1,...,L$ where $x_l$ corresponds to the original sparse vector and $y_l$ corresponds to the observation vector derived from the linear model in (\ref{e3}) for some fixed forward operator $A$ and some unknown forward operator perturbation $E$, that may vary from example to example.
Therefore, by fixing $A$ across training examples and letting $E$ and $n$ vary across training examples, we learn a network that is able to recover $x$ from $y = (A+E) x + n$ for any unknown model perturbation $E$.

We rely on a standard cost function measuring the difference between the network prediction of the vector of interest and the ground truth vector of interest as follows:
\begin{align}
\label{e12}
\min_{\theta} \frac{1}{L} \sum_{l=1}^{L} \|\hat x^l - x^{l} \|^2_2
\end{align}
where $\hat{x}^l$ corresponds to the network output given network input $y^l$, $x^l$ corresponds to the ground truth and $\theta$ aggregates the various learnable parameters.
We then adopt standard gradient descent algorithms in order to learn the network parameters $\lambda, \mu, A_1, A_2$ and $A_3$.

\subsection{REST vs LISTA}
\begin{figure}[h]
\centering
\includegraphics[width=0.59\textwidth]{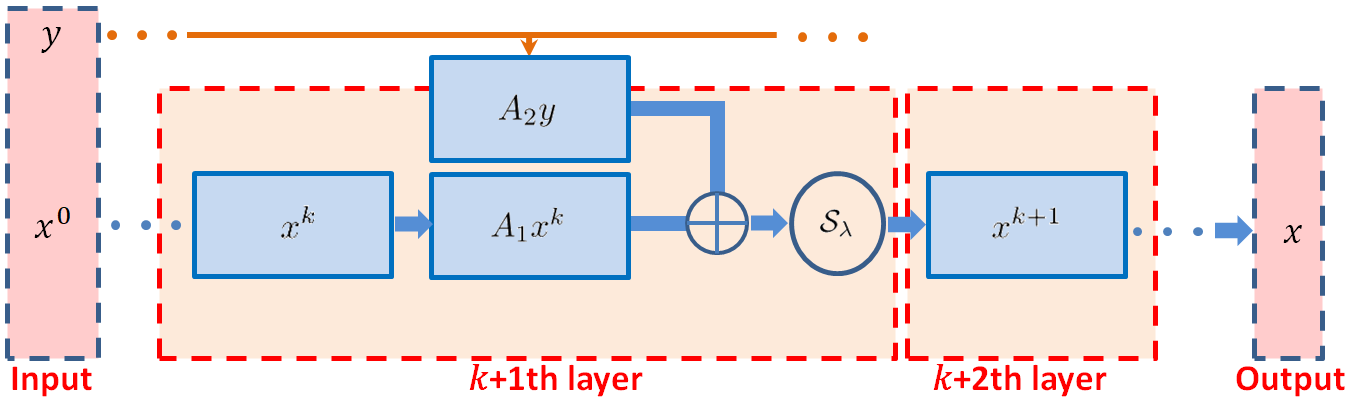}
\caption{LISTA Neural Network Architecture.}\label{fig2}
\end{figure}

We show in sequel that REST can perform substantially better than other unfolding approaches such as LISTA in the presence of inverse problems subject to model uncertainty. It is therefore instructive to compare the REST network architecture to the LISTA architecture (cf. Fig. \ref{fig1} vs Fig. \ref{fig2}).
The main difference between a layer in LISTA and in REST lies in two additional processing operations: 1) one corresponds to the second term in (\ref{e11}) and 2) the other corresponds to the normalization by $1 + \|x\|^2_2$. In fact, without these operations, the REST architecture immediately reduces to a network very similar to LISTA. These additional operations play a critical role in mitigation of sample-wise model mismatch. The first operation enlarges the number of parameters, allowing for a much better fit. In REST, we have $2M \times N + N \times N + 2$ learnable parameters in each layer and comparatively, we have $M \times N + N \times N + 1$ learnable parameters for LISTA.
The second operation normalizes the output of each layer, promoting a more robust solution.


We perform various experiments in Appendix C to support our intuitions on the comparison between REST and LISTA.

\section{Numerical Results}
\label{sec:evaluation}

We now conduct a series of experiments on a simple compressive sensing task where one wishes to recover a vector of interest from a noisy linear measurements, in the presence of model mismatch. We compare the performance of the proposed REST network to well-known LISTA and ADMM-Net as well as results for Basis Pursuit (BP)\cite{BP}, ISTA\cite{ISTA}, robust ISTA\cite{PCS3}, and MU-GAMP\cite{PCS4} algorithms.

\subsection{Experimental setup}

Our experimental set-up involves the generation of synthetic data as follows:
We generate 1100 different sample pairs $(x,y)$ where the matrix is $A$ remains fixed, the matrix $E$ varies from sample to sample, and the noise vector $e$ and target vector $x$ also vary from sample to sample. Up to 1000 of these sample pairs is used for training purposes for the learning-based approaches, i.e., REST, LISTA and ADMM-Net, and the remaining 100 samples are used for testing purposes (i.e. evaluation of the performance of the approaches) for all the different methods. The measurement vector $y \in \mathbb{R}^{13}$ is generated per (\ref{e3}), the target vector $x \in \mathbb{R}^{250}$ is sparse with four randomly chosen non-zero elements taking values uniformly in the interval (0,1], and the noise vector $e \in \mathbb{R}^{130}$ has i.i.d. Gaussian entries with zero mean and variance $\sigma^2=0.03$.
The matrix $A \in \mathbb{R}^{130 \times 250}$ has i.i.d. Gaussian entries with zero mean and variance such that $\|A\|_F = 10$. The matrix $E \in \mathbb{R}^{130 \times 250}$ also has i.i.d. Gaussian entries with zero mean; furthermore, for samples in the training set we generate perturbation matrices $E$ such that $\|E\|_F \leq r$ and for samples in the testing set we generate perturbation matrices $E$ such that $\|E\|_F \leq r'$. This allows us to compare the performance of different learning-based approaches in scenarios where (a) there is no model mismatch in the training samples, in the testing samples, or across training and testing samples ($r = r' = 0$) and (b) there is model mismatch in training samples, testing samples, or across training and testing samples ($r \neq 0$ or $r' \neq 0$).

We use the reconstruction mean squared error (MSE) to evaluate the performance of the different approaches.
\begin{align}
\label{e12}
{\rm MSE} = \sum_{l=1}^L  \frac{1}{LN} \| x^{l} - \hat x^{l}\|^2_2,
\end{align}
where $\hat{x}^l$ corresponds to the network output given network input $y^l$, $x^l$ corresponds to the ground truth.

We consider the various networks listed in Table \ref{T0}.
The learnable parameters of these networks in each layer are tied with shared parameters in each layer.
Note that REST6, ADMM-Net8 and LISTA8 have similar parameter numbers, i.e., 765,012 parameters for REST6, 760,008 parameters for LISTA8 and 760,016 parameters for ADMM-Net8. However, the numbers of learnable parameters are different, i.e., 127,502 learnable parameters for REST6, 95,001 learnable parameters for LISTA8 and 95,002 learnable parameters for ADMM-Net8 as each layer share the same parameters.
\renewcommand{\arraystretch}{1.0}
\begin{table}[tp]
  \centering
  \fontsize{6.5}{6}\selectfont
  \begin{threeparttable}
  \caption{Networks used in Experiments.}\label{T0}
    \begin{tabular}{ll}
    \toprule
    REST6&REST with 6 layers\cr
    LISTA6&LISTA with 6 layers\cr
    LISTA8&LISTA with 6 layers\cr
    ADMM-Net6&ADMM-Net with 6 layers\cr
    ADMM-Net8&ADMM-Net with 8 layers\cr
    \bottomrule
    \end{tabular}
    \end{threeparttable}
\end{table}
We then train the various networks by using stochastic gradient descent (SGD) with learning rate $l_r = 10^{-4}$ and batch size to be 32.
We use 2000 epochs to train the networks.

\subsection{Experimental Results}
\begin{figure*}[t]
    \centering
    \subfigure[]{\label{fig2a}\includegraphics[width=0.325\textwidth]{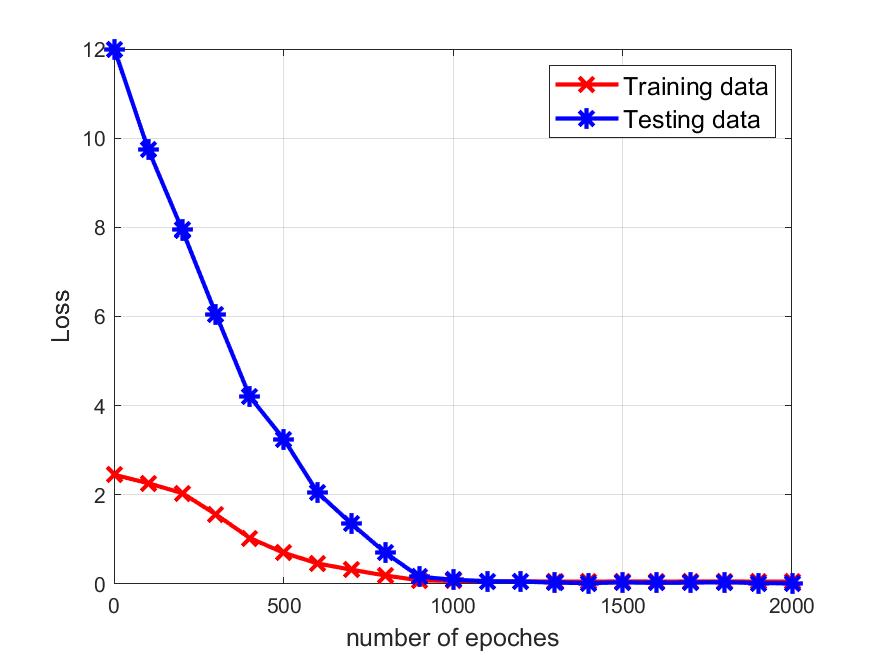}}
        \hfil
    \subfigure[]{\label{fig2b}\includegraphics[width=0.325\textwidth]{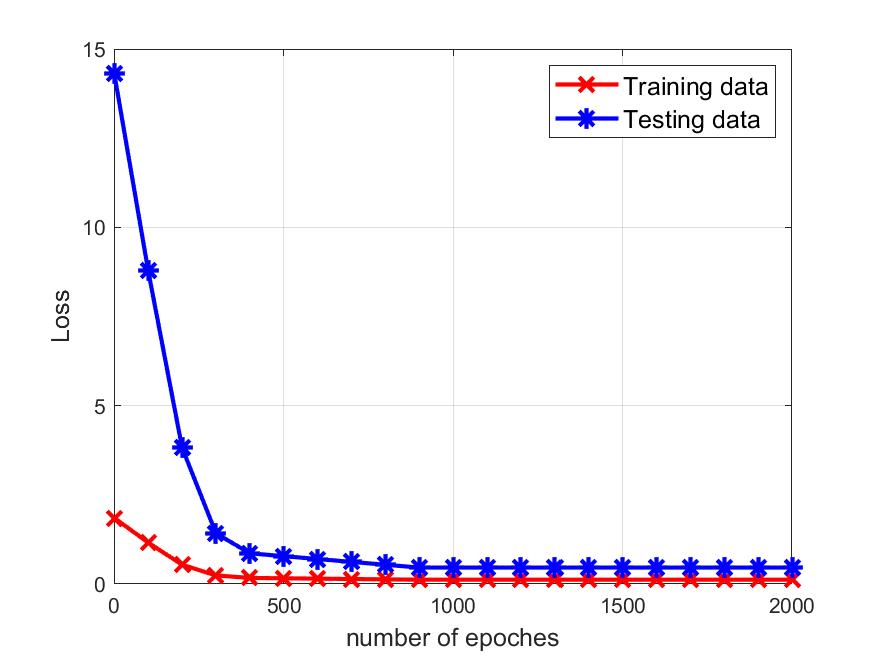}}
        \hfil
    \subfigure[]{\label{fig2b}\includegraphics[width=0.325\textwidth]{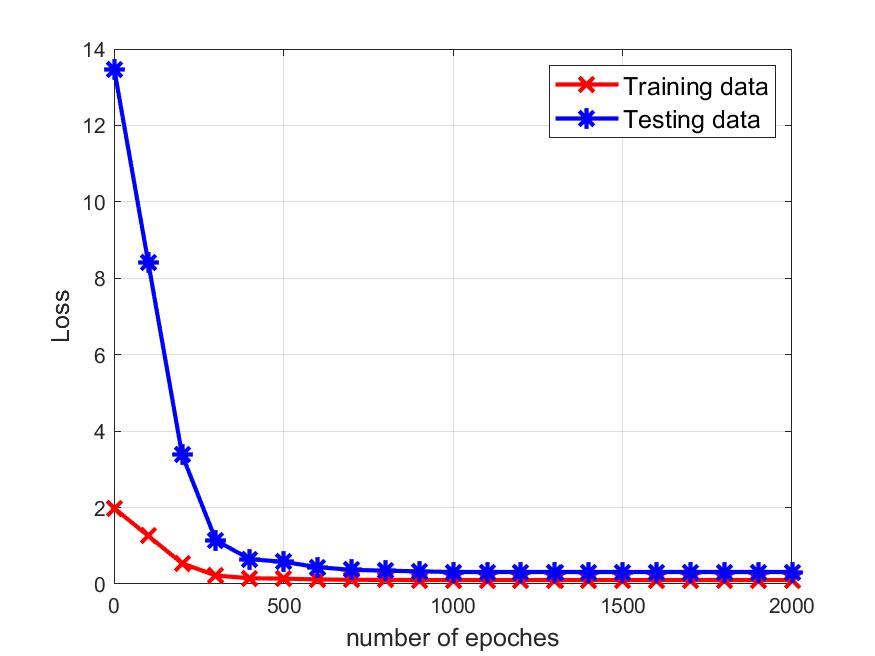}}
        \hfil
    \subfigure[]{\label{fig2b}\includegraphics[width=0.325\textwidth]{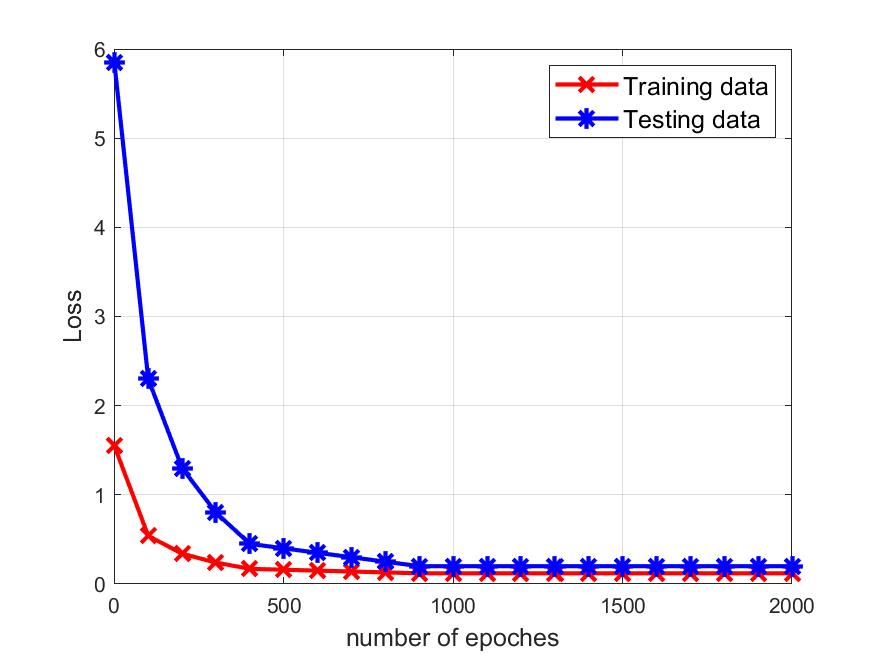}}
        \hfil
    \subfigure[]{\label{fig2b}\includegraphics[width=0.325\textwidth]{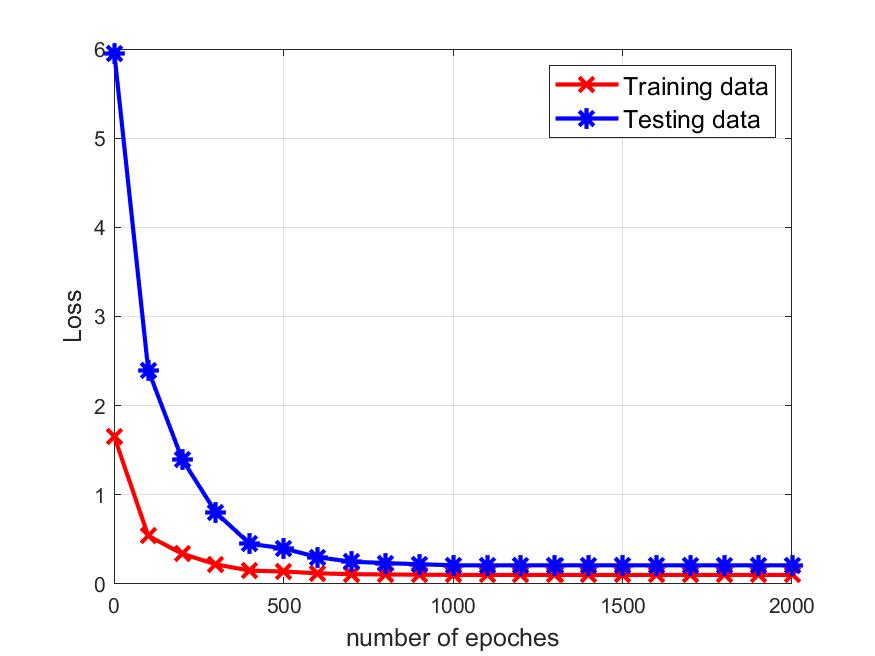}}
    \caption{Training and testing loss vs number of epochs for $r= r' = 5$ with 1000 training samples and 100 testing samples. (a) REST6. (b). LISTA6. (c)LISTA8. (d). ADMM-Net6. (e)ADMM-Net8. }\label{fig3}
\end{figure*}

Fig. \ref{fig3} depicts the training and testing loss in terms of the number of training epochs for the different networks shown in Table \ref{T0} for $r=r'=5$ with 1000 training samples and 100 testing samples.
We observe that LISTA and ADMM-Net converge faster than REST. However, REST fits the data much better than LISTA and ADMM-Net in the presence of model mismatch. In addition, the testing error in REST is lower than LISTA and ADMM-Net, and the generalization error - corresponding to the difference between testing and training errors - of REST is smaller than that of LISTA and ADMM-Net as the number of epochs increase.
Notably, these trends also apply to other values of $r$ and $r'$.

\begin{figure*}[t]
    \centering
    \subfigure[]{\label{fig2a}\includegraphics[width=0.325\textwidth]{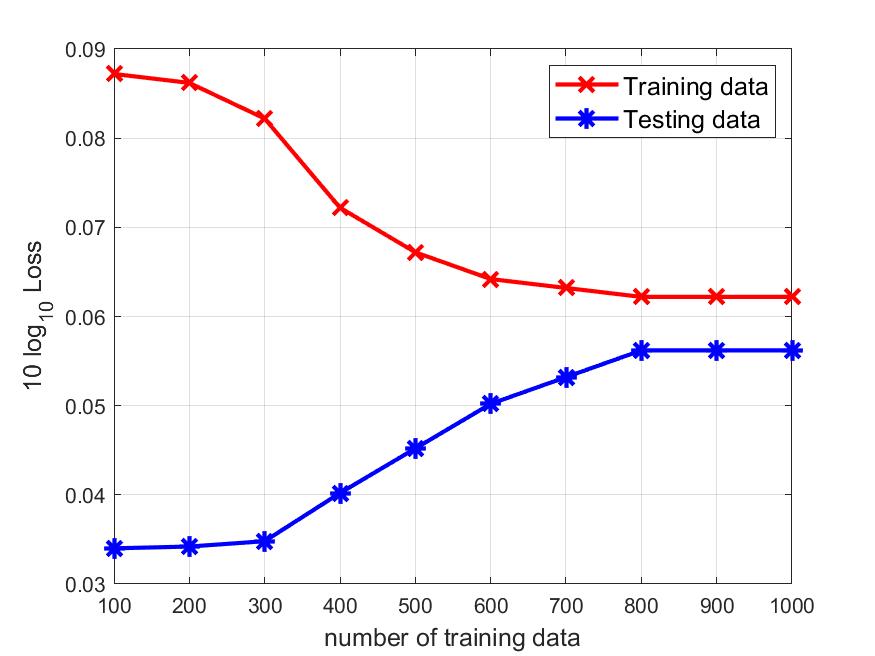}}
        \hfil
    \subfigure[]{\label{fig2b}\includegraphics[width=0.325\textwidth]{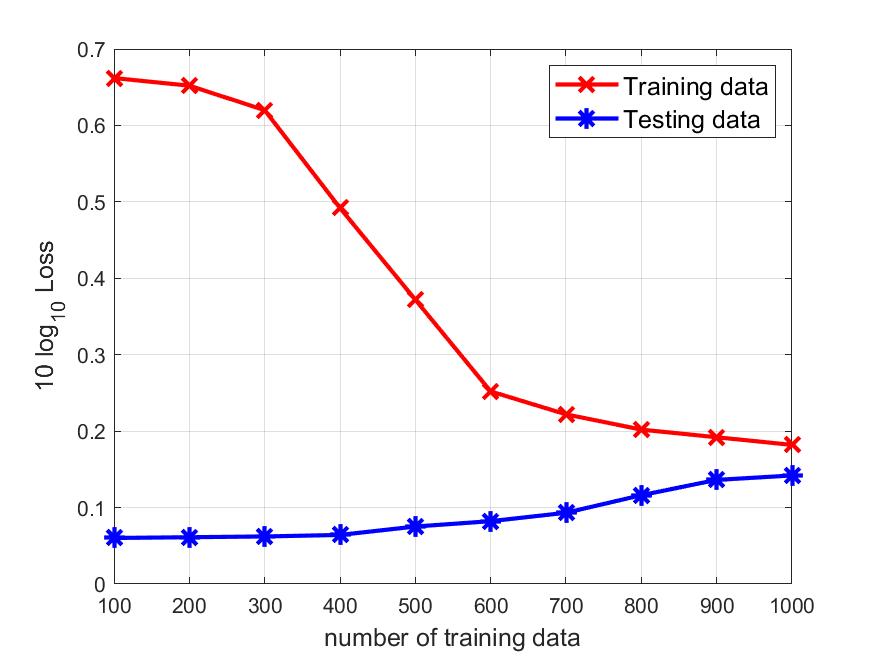}}
        \hfil
    \subfigure[]{\label{fig2b}\includegraphics[width=0.325\textwidth]{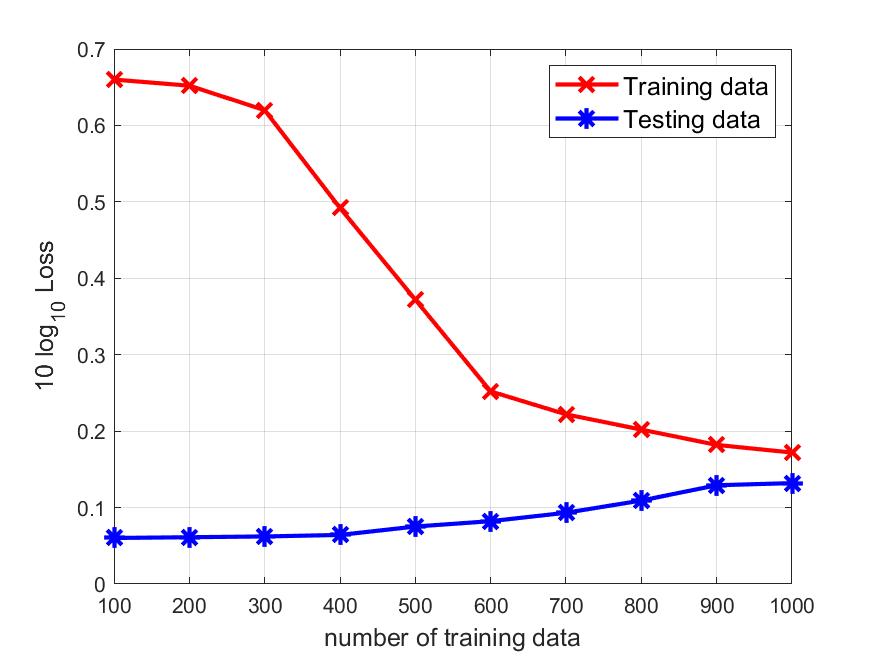}}
        \hfil
    \subfigure[]{\label{fig2b}\includegraphics[width=0.325\textwidth]{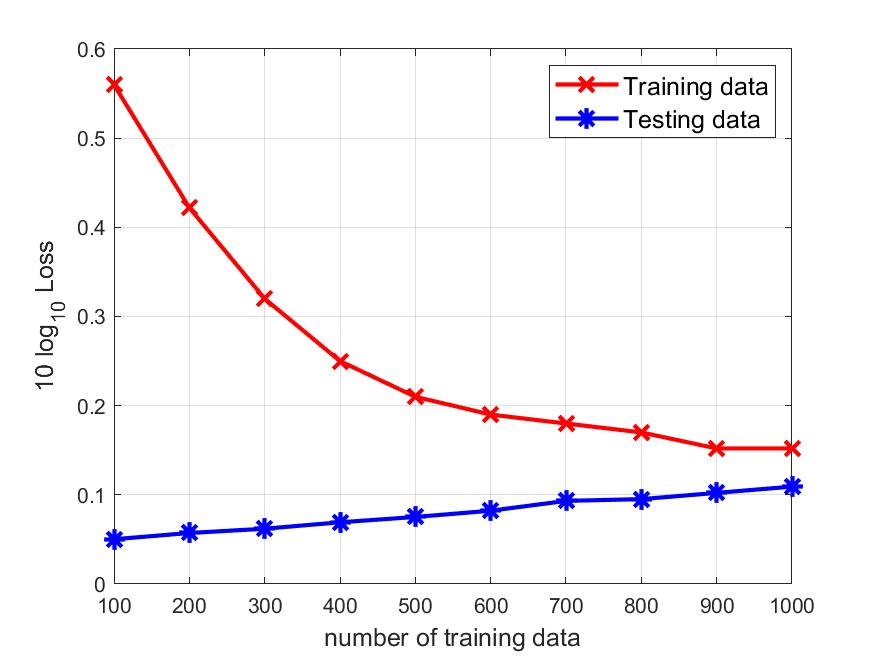}}
        \hfil
    \subfigure[]{\label{fig2b}\includegraphics[width=0.325\textwidth]{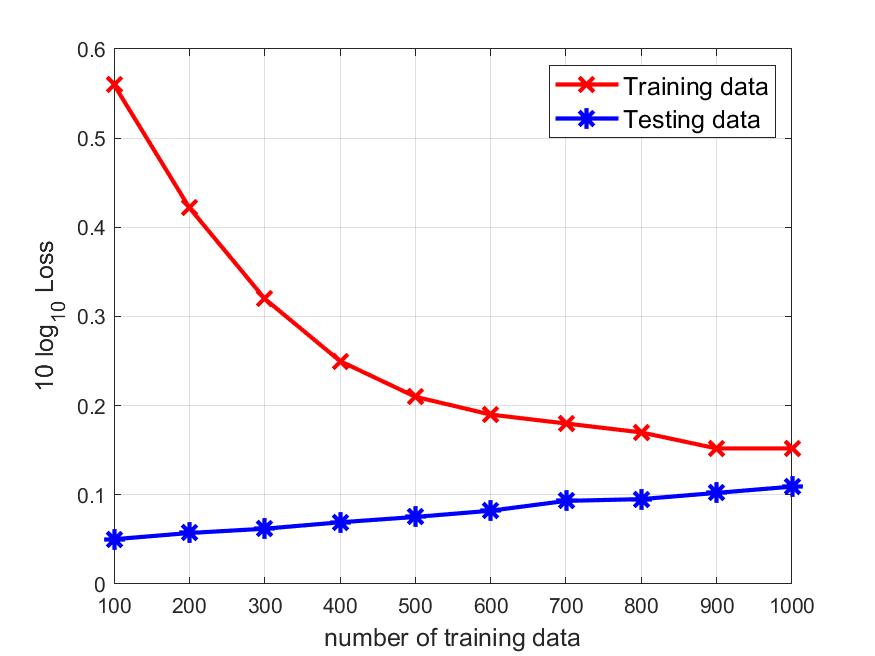}}
    \caption{Training and testing loss vs training set size with 100 testing samples and 2000 epochs $r= r' = 5$ with 2000 epochs. (a) REST6. (b). LISTA6. (c)LISTA8. (d). ADMM-Net6. (e)ADMM-Net8. }\label{fig3.1}
    \end{figure*}

Fig. \ref{fig3.1} depicts training and testing loss vs number of training samples of the different networks shown in Table I for $r=r'=5$ with 100 testing samples and 2000 epochs. We note again that training error is much lower for REST than LISTA and ADMM-Net, especially for a larger number of training points. This indicates that REST fits the training data much better than LISTA and ADMM-Net.
Likewise, the testing error is also much lower for REST in comparison with LISTA and ADMM-Net.
Fig. \ref{fig3.1} also shows that in the presence of model mismatch, the generalization error is smaller for REST than LISTA and ADMM-Net. Similar trends also apply for different values of $r$ and $r'$.

We next evaluate the performance of different approaches as a function of the degree of model mismatch present in training data ($r$) and testing data ($r'$).
Table \ref{T1} depicts training errors ($10\log_{10}$MSE) of REST, LISTA and ADMM-Net for different mismatch errors existing in the training data with 1000 training samples and 2000 epochs.
It can be observed that when the $r = 0$ (i.e. when the forward model is fixed so that each training sample sees the same forward model), LISTA and ADMM-Net have smaller training errors than REST. This is because LISTA and ADMM-Net apply to inverse problems without any model mismatch.
However, when the $r \ge 2$, LISTA and ADMM-Net do not fit well to the training data whereas REST does much better.

\renewcommand{\arraystretch}{1.2}
\begin{table*}[tp]
  \centering
  \fontsize{6.5}{6}\selectfont
  \begin{threeparttable}
  \caption{Training loss ($10\log_{10}$MSE) of different networks with different $r$ with 1000 training samples and 2000 epochs.}\label{T1}
    \begin{tabular}{ccccccccccccc}
    \toprule
    &$r = 0$&$r = 2$&$r = 4$&$r = 6$&$r = 8$&$r = 10$&$r = 12$\cr
    \midrule
    REST6&-2.61&-1.98&-1.67&-1.28&-1.06&-0.94&-0.81\cr
    LISTA6&-3.70&-1.14&-0.97&-0.61&-0.34&0.29&0.45\cr
    LISTA8&-3.74&-1.15&-1.02&-0.66&-0.36&0.28&0.44\cr
    ADMM-Net6&-3.96&-1.57&-1.01&-0.70&-0.48&0.12&0.35\cr
    ADMM-Net8&-3.98&-1.57&-1.05&-0.73&-0.49&0.11&0.33\cr
    \bottomrule
    \end{tabular}
    \end{threeparttable}
\end{table*}

Table \ref{T2} depicts the testing errors ($10\log_{10}$MSE) of different networks with different $r$ and $r'$ with 1000 training samples, 100 testing samples and 2000 epochs.
We now comment on the performance under different mismatch regimes:
\begin{itemize}
\item Scenario $r = r' = 0$: This scenario is such that the training samples and the testing samples are subject to the same forward operator. It is clear that LISTA and ADMM-Net have better performance than REST because LISTA and ADMM-Net are specifically designed for inverse problems without any mismatch.

\item Scenario $r \neq 0$ or $r' \neq 0$: This scenario is such that the training samples and/or the testing samples are subject to different forward operators. We can observe that -- for a certain level of mismatch in the testing samples $r' > 0$ -- the higher the amount of mismatch in the training set, the higher the performance of the various unrolling approaches, i.e. REST, LISTA, and ADMM-Net. This is due to the fact that optimizing any of the networks with training data that is subject to different models inherently leads to a more robust network (i.e. this is an instance of robust training). We also observe that for a certain level of mismatch in the training samples $r > 0$ -- the higher the amount of model mismatch in the testing samples the lower the performance of the various networks. Importantly, both in scenarios where $r > r'$, $r < r'$, or $r=r' \neq 0$, the proposed REST network outperforms any of the competing networks.
\end{itemize}

\renewcommand{\arraystretch}{1.2}
\begin{table*}[t]

  \centering
  \fontsize{6.5}{6}\selectfont
  \begin{threeparttable}
  \caption{Testing loss ($10\log_{10}$MSE) for different networks with different $r$ and $r'$ with 1000 training samples, 100 testing samples and 2000 epochs.}\label{T2}
    \begin{tabular}{ccccccccccccc}
    \toprule
    \multirow{3}{*}{Methods}&
    \multicolumn{4}{c}{ $r = 0$}&\multicolumn{4}{c}{$r = 4$}&\multicolumn{4}{c}{$r= 8$}\cr
    \cmidrule(lr){2-5} \cmidrule(lr){6-9} \cmidrule(lr){10-13}
    &$r' = 0$&$r' = 4$&$r' = 8$&$r' = 12$&$r' = 0$&$r' = 4$&$r' = 8$&$r' = 12$&$r' = 0$&$r' = 4$&$r' = 8$&$r' = 12$\cr
    \midrule
    REST6&-2.57&-1.58&-0.99&-0.74&-2.40&-1.72&-1.22&-1.01&-2.05&-1.87&-1.52&-1.25\cr
    LISTA6&-3.49&-0.62&0.48&0.54&-2.57&-0.74&-0.26&0.09&-1.60&-0.78&-0.47&-0.22\cr
    LISTA8&-3.49&-0.65&0.47&0.52&-2.56&-0.79&-0.27&0.07&-1.62&-0.81&-0.53&-0.27\cr
    ADMM-Net6&-3.77&-0.53&0.62&0.81&-2.91&-1.03&-0.48&-0.13&-1.77&-1.01&-0.66&-0.40\cr
    ADMM-Net8&-3.78&-0.54&0.61&0.80&-2.90&-1.05&-0.51&-0.16&-1.77&-1.01&-0.69&-0.41\cr
    \bottomrule
    \end{tabular}
    \end{threeparttable}
\end{table*}

Table \ref{T3} also depicts the testing errors ($10\log_{10}$MSE) of different model-based approaches.
\renewcommand{\arraystretch}{1.2}
\begin{table}[t]
  \centering
  \fontsize{6.5}{6}\selectfont
  \begin{threeparttable}
  \caption{Errors ($10\log_{10}$MSE) of testing data for different model-based approaches with different $r'$..}\label{T3}
    \begin{tabular}{ccccccccccccc}
    \toprule
    &$r' = 0$&$r' = 4$&$r' = 8$&$r' = 12$\cr
    \midrule
    BP&-3.07&-1.26&0.45&0.91\cr
    ISTA&-2.82&-0.73&0.43&0.83\cr
    Robust ISTA&-2.47&-1.22&-0.48&-0.17\cr
    MU-GAMP&-2.39&-1.30&-0.99&-0.70\cr
    \bottomrule
    \end{tabular}
    \end{threeparttable}
\end{table}
As expected, comparing Table III to Table IV, without any model mismatch in the testing data ($r'=0$), model-based approaches outperform REST (independently of the value of $r$); conversely, with model mismatch in the testing data ($r'>0$), REST outperforms any of the model-based approaches, for any value of $r$ (the level of model mismatch in the training data).

\begin{figure}[h]
\centering
\includegraphics[width=0.325\textwidth]{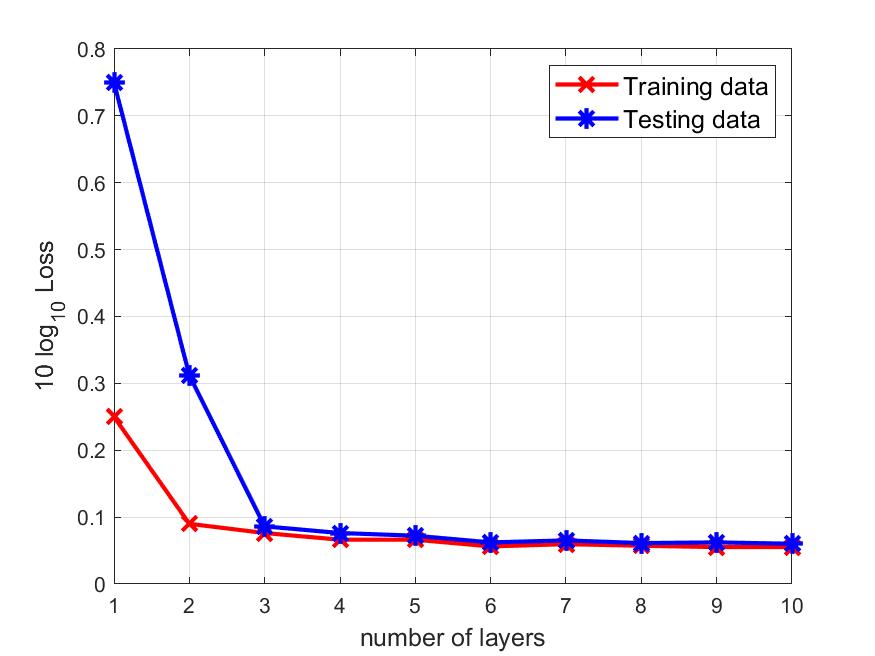}
\caption{Performance of REST with different layer numbers with 1000 training samples, 100 testing samples and 2000 epochs}\label{fig4}
\end{figure}
We also evaluate the performance of the proposed REST network as a function of the number of layers. We use the aforementioned default experiment setting except that we now vary the  number of layers from 1 to 10. It is clear from Fig. \ref{fig4} that the number of layers does not appear to affect significantly the training loss. In contrast, it is also clear that the number of layers can affect the testing loss. We attribute this observation to the fact that unfolding networks architecture comes with a strong inductive bias, so that training data can be well fitted with a small number of layers. However, when the number of network layers is small, the processing capacity of REST for complex data is poor. Although REST with a small number of layers can fit training data well, the performance is poor when it comes to the new data during testing procedure.

Finally, we evaluate the computational cost of the different approaches. The different algorithms were implemented on a work server with 4 NVIDIA Tesla V100 GPU. The running times (seconds) to train the networks REST6, LISTA6, LISTA8, ADMM-Net6 and ADMM-Net8 are 41.7s, 29.8s, 39.4s, 26.2s and 37.1s, respectively, while the running times of using the learnt networks to predict the results on testing data of REST6, LISTA6, LISTA8, ADMM-Net6 and ADMM-Net8 are 3.0$\times10^{-4}$s, 7.8$\times10^{-5}$s, 1.0$\times10^{-4}$s, 7.9$\times10^{-5}$s and 1.0$\times10^{-4}$s, respectively. The running time of BP, ISTA, robust ISTA and MU-GAMP algorithms are 8.3$\times10^{-3}$s, 8.9$\times10^{-3}$s, 2.2$\times10^{-2}$s and 3.7$\times10^{-2}$s, respectively. Although the computational cost of the training phase is relatively high, optimized networks deliver estimation results much faster than the model-based approaches. Notably, training complexity of REST is not considerably higher than that of LISTA or ADMM-Net, with the slightly higher complexity offset by much better performance.

\section{REST in radar imaging}
We finally apply the proposed REST network in a synthetic aperture radar (SAR) imaging application subject to model mismatch.

\subsection{Problem formulation}

\begin{figure}[t]
\centering
\includegraphics[width=0.43\textwidth]{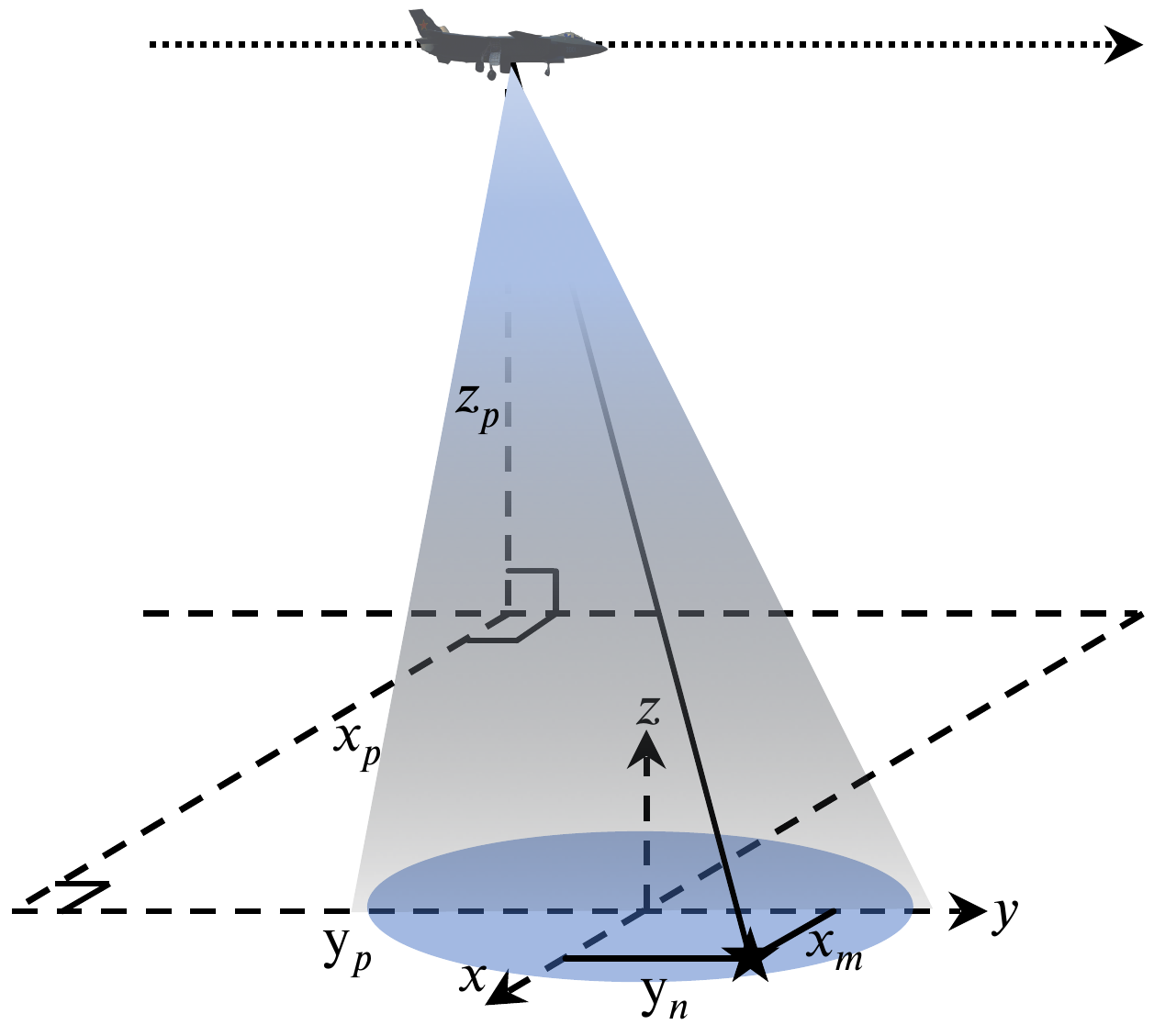}
\caption{Configuration of SAR system.}\label{figG}
\end{figure}

We consider a SAR imaging problem where the radar transmits linear frequency modulated (LFM) pulses at a constant rate. Let $x_m$ and $y_n$ refer to the $m$th range gate and $n$th azimuth bin of the target SAR image $Z$, respectively, where $m=1,2,\cdots,M$ and $n=1,2,\cdots,N$. $M$ and $N$ represent the sample numbers of range and azimuth direction, respectively. It is assumed that $\tau$ and $t$ respectively represent the range and azimuth time, and $i=1,2,\cdots,I$ and $q=1,2,\cdots,Q$ index the range and azimuth time, where $I$ and $Q$ indicate the sample number of range and azimuth time, respectively. Set $s(\tau_i, t_q,x_m,y_n)$ as the received signal of the $q$th pulse in the $i$th sequence from the target located at $(x_m,y_n)$, which can be written as follows:
\begin{equation}
\label{1.1}
s(\tau_i,t_q,x_m,y_n)=\varphi(\tau_i,t_q,x_m,y_n) {{Z}}(x_m,y_n),
\end{equation}
where ${{Z}}(x_m,y_n)$ denotes the reflector coefficient of the target located at $(x_m,y_n)$, and
\begin{align}
\label{1.2}
\varphi(\tau_i,t_q,x_m,y_n)=&
{\omega _r}\left( {{\tau_i}} \right){\omega _a}\left( {{t_q} } \right)\nonumber\\
&\cdot\exp \left\{ { - j\pi {k_r}{{\left( {{\tau_i} - \frac{{{R}({t_q},x_m,y_n)}}{c}} \right)}^2}} \right\}\nonumber\\
&\cdot\exp \left\{ { - j4\pi {f_c}\frac{{{R}({t_q},x_m,y_n)}}{c}} \right\}.
\end{align}
In (\ref{1.2}), $j = \sqrt{-1}$, $f_c$ is the carrier frequency, $c$ is the speed of light, $k_r$ indicates the frequency modulation (FM) rate associated with the transmitted LFM signal, ${\omega _r}$ and ${\omega _a}$ denote range and azimuth envelope corresponding  to rectangular window functions given by
\begin{align}
\label{5.1}
{\omega _r}(\tau_i)= \left\{ {\begin{array}{*{20}{c}}
1, \quad {\rm if} \quad |\tau_i-{2{R}({t_q},x_m,y_n)}/{c}|< \frac{T_r}{2},\\
0, \quad {\rm otherwise},
\end{array}} \right.
\end{align}
and
\begin{align}
\label{5.2}
{\omega _a}(t_q)= \left\{ {\begin{array}{*{20}{c}}
1,\quad {\rm if} \quad |t_q-{y_n}/{v}|< \frac{T_s}{2},\\
0, \quad {\rm otherwise}.
\end{array}} \right.
\end{align}
where $T_r$ and $T_s$ denote the transmitted signal time width and synthetic aperture time, respectively, and
${{R}({t_q},x_m,y_n)}$ represents the instantaneous slant range from image position $(x_m,y_n)$ to the moving platform at azimuth time $t_q$
\begin{align}
\label{5.5}
{R}({t_q},x_m,y_n)=\sqrt {(y_p+vt_q-y_n)^2+(x_p-x_m)^2+z_p^2},
\end{align}
where $(x_p, y_p, z_p)$ denotes radar platform position at the time $t_q =0$, and the radar platform is assumed to move along $y$ axis with constant velocity $v$. The corresponding SAR system configuration is illustrated in Fig. \ref{figG}.

Assume the received signal at the $q$th pulse of the $i$th sequence is represented by ${{S}}(\tau_i,t_q)$, which can be written as follows:
\begin{align}
\label{1.3}
{{S}}(\tau_i,t_q) & = \sum\limits_m {\sum\limits_n {s(\tau_i,t_q,x_m,y_n)} } + \Gamma(\tau_i,t_q) \nonumber\\
& = \sum\limits_m {\sum\limits_n {\varphi(\tau_i,t_q,x_m,y_n) Z(x_m,y_n)} } + \Gamma(\tau_i,t_q),
\end{align}
where $\Gamma (\tau_i,t_q)$ denotes the additive noise at $(\tau_i,t_q)$.
The equation can be also rewritten as
\begin{align}
\label{1.4}
y = A x + e,
\end{align}
where $y = {\rm vec } ({{S}})$, $x = {\rm vec } ({{Z}})$, $e = {\rm vec } (\Gamma)$, $H$ is
\begin{align}
\label{1.5}
A = \left[ {\begin{array}{*{20}{c}}
\varphi_{1,1,1,1} &\varphi_{1,1,2,1} &\cdots&\varphi_{1,1,M,N}\\
\varphi_{2,1,1,1} &\varphi_{2,1,2,1} &\cdots&\varphi_{2,1,M,N}\\
\vdots&\vdots&\ddots&\vdots\\
\varphi_{I,Q,1,1} &\varphi_{I,Q,2,1} &\cdots& \varphi_{I,Q,M,N}\\
\end{array}} \right] ,
\end{align}
and $\varphi_{i,q,m,n} = \varphi(\tau_i,t_q,x_m,y_n)$.
Subsequently, the radar imaging problem can be cast onto the problem:
\begin{align}
\label{1.6}
\min_x & \quad \| y - Ax\|_2 + \lambda \| x \|_1.
\end{align}

In the practical application of SAR imaging, because of influence the of platform vibration, wind field, and turbulence, the antenna phase center of the radar platform might significantly deviate from the ideal trajectory, thus leading to errors in terms of instantaneous slant range ${{R}({t_q},x_m,y_n)}$. Errors in ${{R}({t_q},x_m,y_n)}$ introduce mismatch in the nominal measurement $A$ appearing in (\ref{1.4}) and (\ref{1.6}). In principle, it is possible to calculate errors of ${{R}({t_q},x_m,y_n)}$ using motion measurement data recorded by an ancillary measurement instrument such as inertial navigation system (INS), inertial measurement units (IMUs), and global positioning system (GPS). However, these measuring instruments are not precise enough for the required accuracy, and in many cases, there are no usable motion measurement data.

Suppose the slant range is
\begin{align}
\label{5.6}
&{R}'({t_q},x_m,y_n)= \nonumber\\
&\sqrt {(y_p+vt_q-y_n- \Delta y_{q})^2+(x_p-x_m - \Delta x_q)^2+(z_p- \Delta z_{q})^2},
\end{align}
where $\Delta x_{q}$, $\Delta y_{q}$ and $\Delta z_{q}$ denote the motion deviations from the nominal trajectory at azimuth time $t_q$ with respect to $x$, $y$ and $z$ directions, respectively.
Replacing ${R}({t_q},x_m,y_n)$ in (\ref{1.2}) and (\ref{1.5}) by ${R}'({t_q},x_m,y_n)$ leads to a measurement model associated with the operator $A+E$ instead of $A$ only, where $E$ captures the difference between ${R}({t_q},x_m,y_n)$ and ${R'}({t_q},x_m,y_n)$. The original model in (\ref{1.4}) then becomes
\begin{align}
\label{1.7}
y = (A+E)x +e,
\end{align}
implying that we can leverage REST for SAR imaging applications.

Note, however, that in SAR imaging problem, the variables in (\ref{1.7}) are complex rather than real. To use the proposed REST network, we decompose the variable into their real and imaginary parts.

\subsection{Experimental Results}
We compare the performance of the proposed REST network to LISTA, ADMM-Net, BP, ISTA and robust ISTA, and MU-GAMP algorithms.

\subsubsection{Experimental setup}
System and geometry parameters are listed in Table \ref{T33}.
\renewcommand{\arraystretch}{1.0}
\begin{table}[htpb]
\centering
\caption{Simulation parameters}\label{T33}
\begin{tabular}{l r}
\hline\hline
 $\mathbf{Parameter}$          & $\mathbf{Value}$\\
 \hline
$f_c$: Carrier frequency              &$10\mathrm{GHz}$\\
$B_r$: Signal bandwidth                     & $400\mathrm{MHz}$\\
$T_r$: Signal time width                     & $1.5\mathrm{\mu s}$\\
$f_s$: Fast time sampling rate       & $500\mathrm{MHz}$\\
$T_s$: Synthetic aperture time         &$2\mathrm{s}$\\
$v$: Radar platform velocity       &$200m/s$\\
$PRF$: Pulse repetition frequency        & $800\mathrm{Hz}$\\
$x_p$: Radar position in $x$ direction at $t =0$          & $3\mathrm{km}$\\
$y_p$: Radar position in $y$ direction at $t =0$    & $0\mathrm{km}$\\
$z_p$: Radar position in $z$ direction at $t =0$    & $5\mathrm{km}$\\
\hline\hline
\end{tabular}
\end{table}
Here, we randomly generate $\Delta x_{q}$, $\Delta y_{q}$ and $\Delta z_{q}$ in (\ref{5.6}) with the constraint $\Delta x_{q} \in [-d,d]$, $\Delta y_{q}\in [-d,d]$ and $\Delta z_{q} \in [-d,d]$. Note that different values of $d$ correspond to different $r$ and $r'$.

\begin{figure}[h]
    \centering
    \subfigure{\label{fig2b}\includegraphics[width=0.35\textwidth]{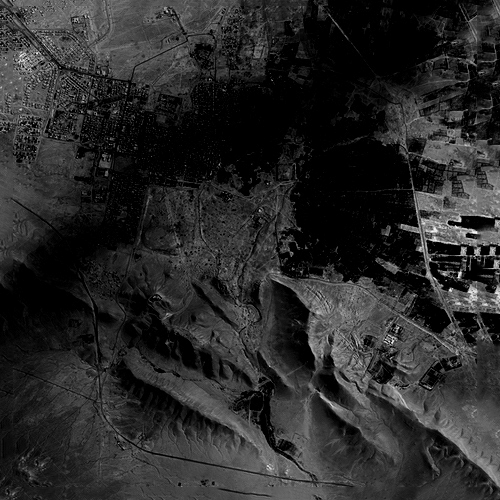}}
    \hfil
    \subfigure{\label{fig2b}\includegraphics[width=0.35\textwidth]{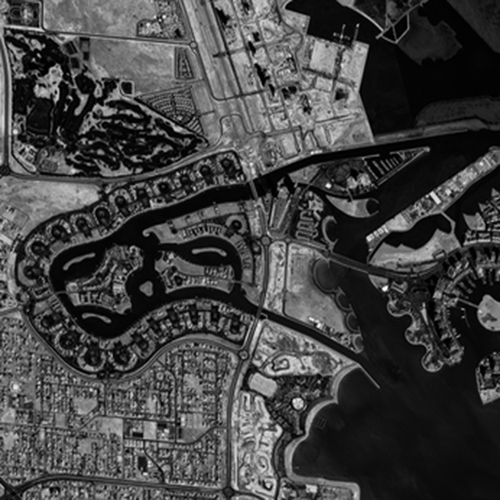}}

    \subfigure{\label{fig2c}\includegraphics[width=0.35\textwidth]{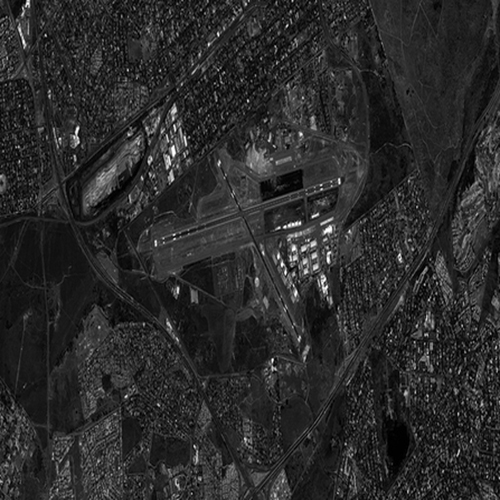}}
    \hfil
    \subfigure{\label{fig2c}\includegraphics[width=0.35\textwidth]{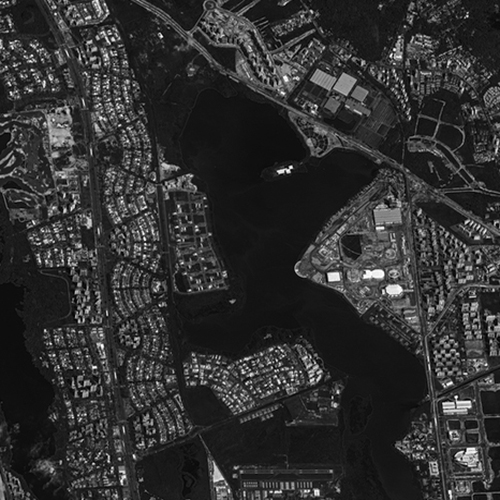}}
    \caption{SAR images used in training process.}\label{figr1}
\end{figure}

\begin{figure}[t]
\centering
\includegraphics[width=0.35\textwidth]{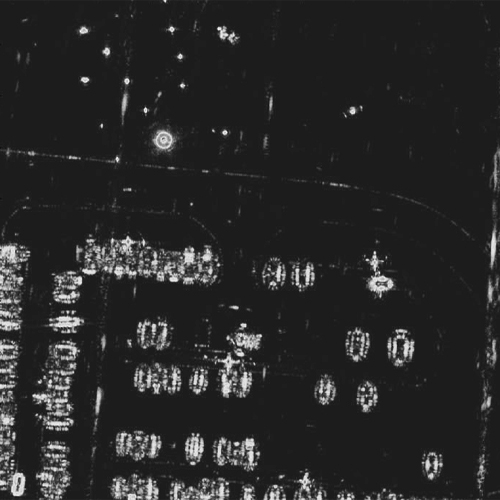}
\caption{SAR image used in testing process.}\label{figr2}
\end{figure}

In the training process, 400 different SAR images are used as reflector coefficient matrix $Z$, which are obtained by dividing each image in Fig. \ref{figr1} (a)-(d) into 100 small images.
As for the testing data, real SAR images are used as reflector coefficient matrix $Z$ by dividing the image in Fig. \ref{figr2} into 256 small images.
We set the size of single data as $M=512$ and $N=128$.
Both real and imagery parts of noise vector $\eta$ have i.i.d. Gaussian entries with zero mean and variance $\sigma^2=0.01$.
Then, values of $y$, $A$, $E$, $x$ and $e$ in radar imaging task can be generated for both training and testing data.

\renewcommand{\arraystretch}{1.2}
\begin{table}[tp]
  \centering
  \fontsize{6.5}{6}\selectfont
  \begin{threeparttable}
  \caption{Definitions of different networks used in Experiments.}\label{T3.0}
    \begin{tabular}{ll}
    \toprule
    REST4&REST with 4 layers\cr
    LISTA4&LISTA with 4 layers\cr
    LISTA7&LISTA with 7 layers\cr
    ADMM-Net4&ADMM-Net with 4 layers\cr
    ADMM-Net7&ADMM-Net with 7 layers\cr
    \bottomrule
    \end{tabular}
    \end{threeparttable}
\end{table}
We consider the various networks listed in Table \ref{T3.0}, wherein the learnable parameters of these networks in each layer are set to be tied. Note that REST4, ADMM-Net7 and LISTA7 have the similar parameter numbers, i.e., 1,572,872 parameters for REST4, 2,293,767 parameters for LISTA7 and 2,293,774 parameters for ADMM-Net7. However, the numbers of learnable parameters are different, i.e., 393,218 learnable parameters for REST4, 327,681 learnable parameters for LISTA7 and 327,682 learnable parameters for ADMM-Net7.

We set two different training approaches to train REST4, LISTA4, LISTA7, ADMM-Net4 and ADMM-Net7:
\begin{itemize}
\item \textbf {Training strategy 1:} When generating training data, we set $d = 0$ (hence $E=0$). In this case, there is no mismatch in the training data. We are interested in assessing how the performance of different designed networks compare in the presence of model mismatch at testing time without any form of training robustization.
\item \textbf {Training strategy 2:} When generating training data, we set $d = 0.16$, and generate 400 different mismatch matrices. Among them, we observe that the maximum value of $\| E\|_F$ is 3.01, and therefore, $r = 3.01$. In this case, we model mismatch in both training and testing data in the sense that mismatch changes from sample to sample, so we are interested in assessing how the performance of different networks compare in the presence of model mismatch at testing time with a robust training approach.
\end{itemize}

The various networks are also trained by using SGD with learning rate $l_r = 4 \times 10^{-4}$ and batch size to be 32.
We use totally 2000 epochs to train the networks.

\subsubsection{Experimental Results}
\begin{figure}[htb]
    \centering
    \subfigure[]{\label{fig2a}\includegraphics[width=0.325\textwidth]{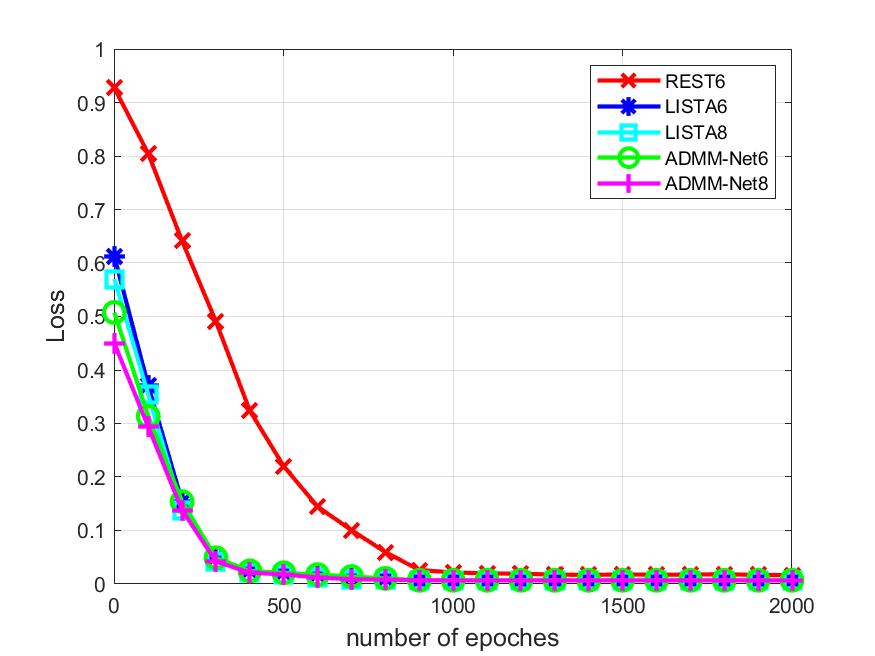}}
    \hfil
    \subfigure[]{\label{fig2b}\includegraphics[width=0.325\textwidth]{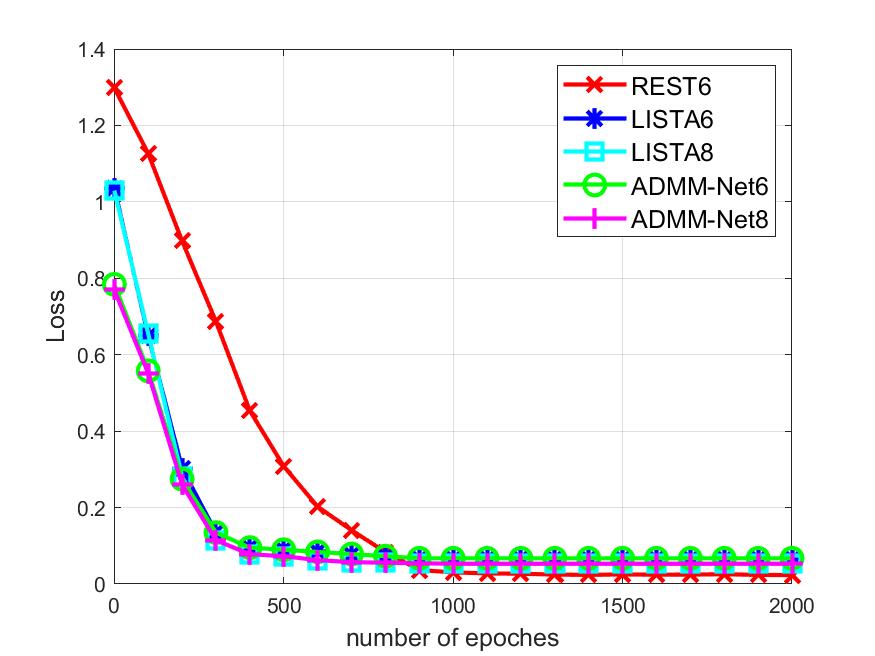}}
    \caption{Training loss vs number of training epochs for different networks. (a). Training strategy 1. (b). Training strategy 2.}\label{figr22}
\end{figure}

The evolution of the training loss in terms of the number of training epochs of different networks under training strategies 1 and 2 are shown in Fig. \ref{figr22} (a) and (b), respectively. It can be observed that in both training strategies 1 and 2, LISTA and ADMM-Net converge faster. However, when the training data contains model mismatch in training strategy 2, REST fits the data much better than LISTA and ADMM-Net.

\begin{figure}[h]
    \centering
    \subfigure{\label{fig2b}\includegraphics[width=0.135\textwidth]{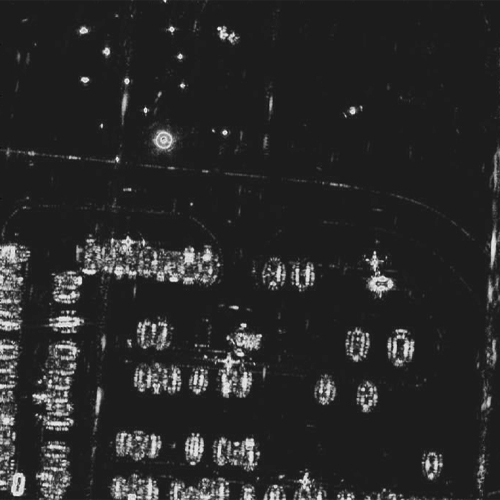}}
    \hfil
    \subfigure{\label{fig2b}\includegraphics[width=0.135\textwidth]{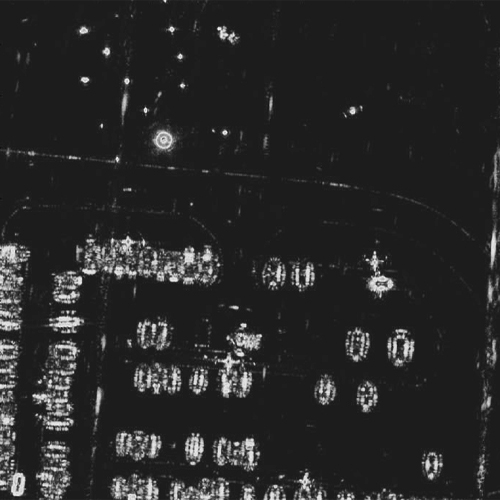}}
    \hfil
    \subfigure{\label{fig2b}\includegraphics[width=0.135\textwidth]{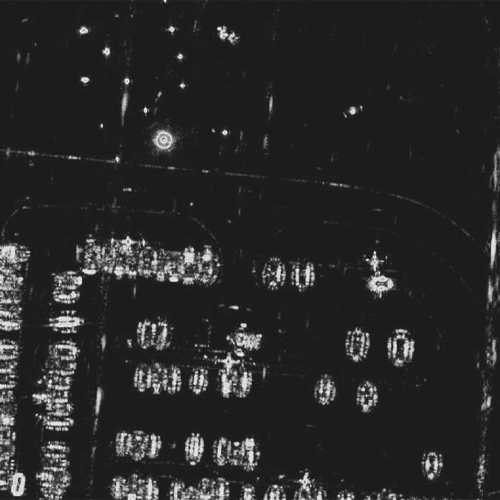}}
    \hfil
    \subfigure{\label{fig2b}\includegraphics[width=0.135\textwidth]{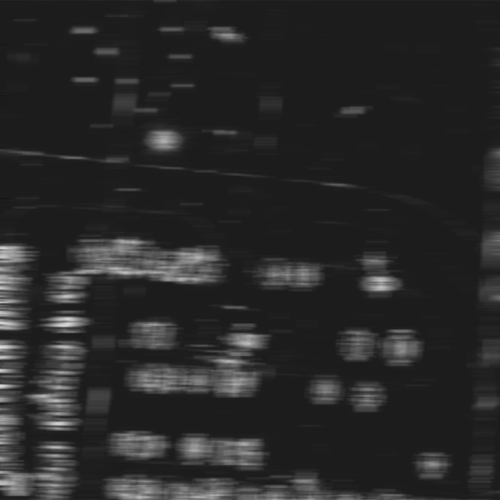}}
    \hfil
    \subfigure{\label{fig2b}\includegraphics[width=0.135\textwidth]{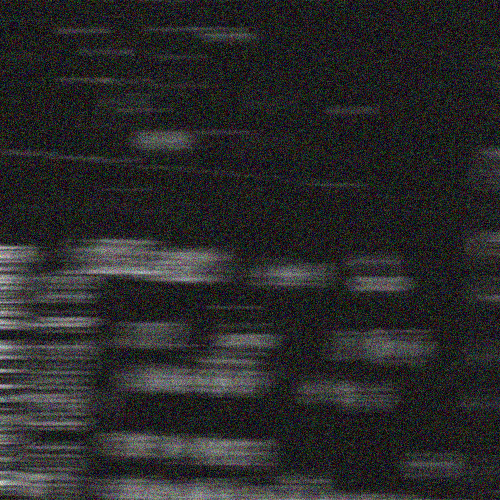}}

    \subfigure{\label{fig2b}\includegraphics[width=0.135\textwidth]{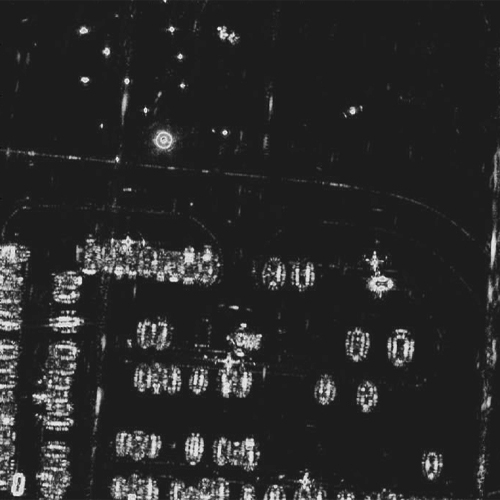}}
    \hfil
    \subfigure{\label{fig2b}\includegraphics[width=0.135\textwidth]{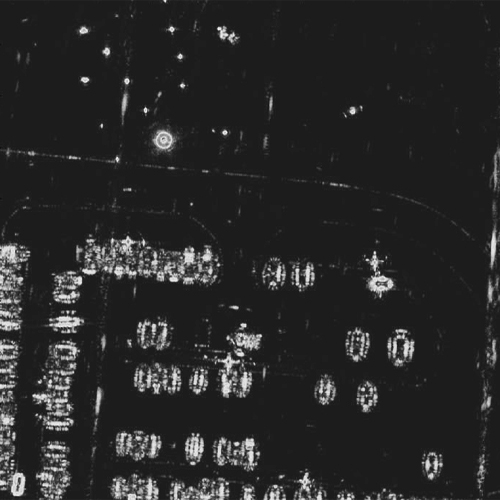}}
    \hfil
    \subfigure{\label{fig2b}\includegraphics[width=0.135\textwidth]{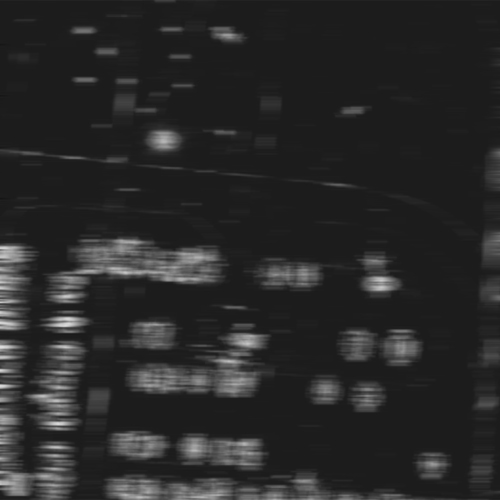}}
    \hfil
    \subfigure{\label{fig2b}\includegraphics[width=0.135\textwidth]{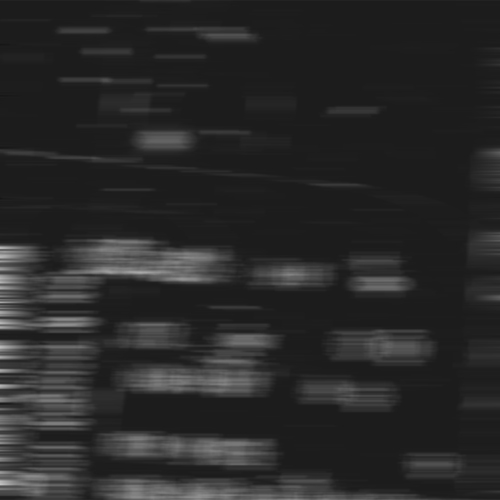}}
    \hfil
    \subfigure{\label{fig2b}\includegraphics[width=0.135\textwidth]{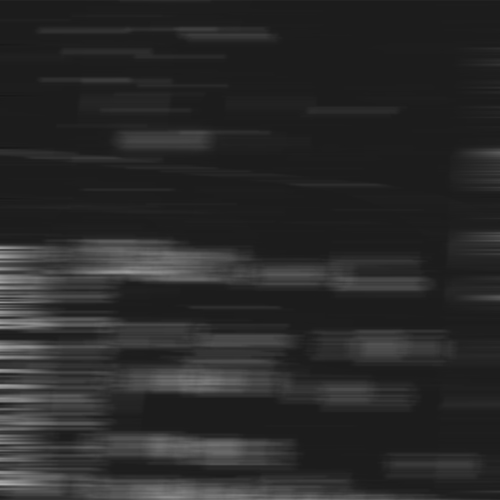}}

    \subfigure{\label{fig2b}\includegraphics[width=0.135\textwidth]{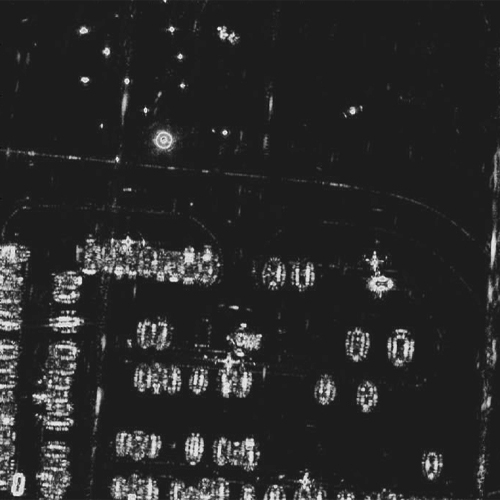}}
    \hfil
    \subfigure{\label{fig2b}\includegraphics[width=0.135\textwidth]{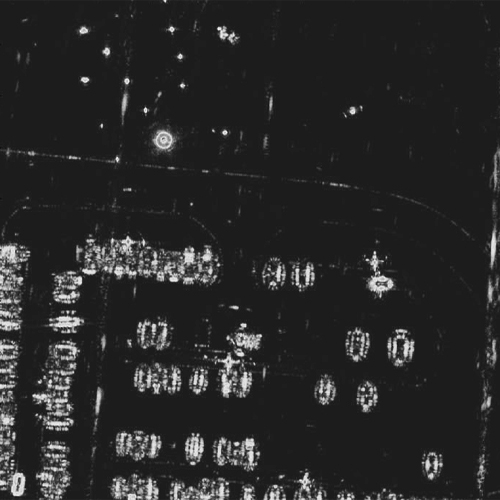}}
    \hfil
    \subfigure{\label{fig2b}\includegraphics[width=0.135\textwidth]{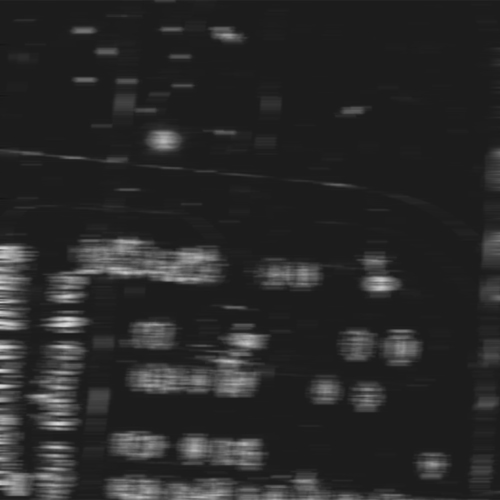}}
    \hfil
    \subfigure{\label{fig2b}\includegraphics[width=0.135\textwidth]{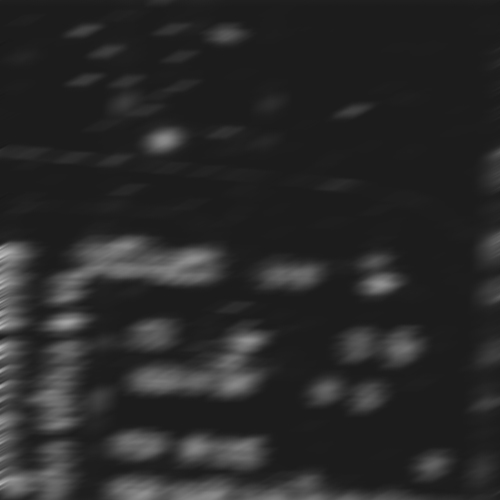}}
    \hfil
    \subfigure{\label{fig2b}\includegraphics[width=0.135\textwidth]{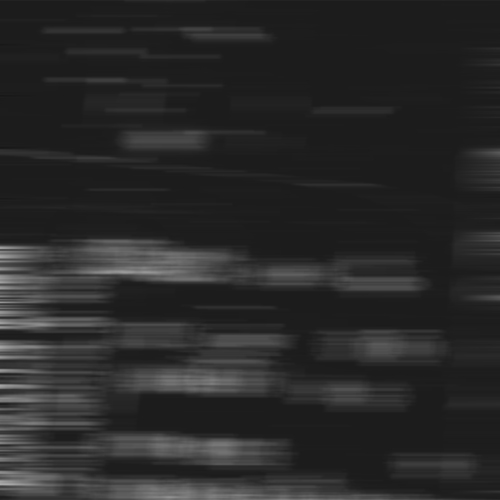}}

    \subfigure{\label{fig2b}\includegraphics[width=0.135\textwidth]{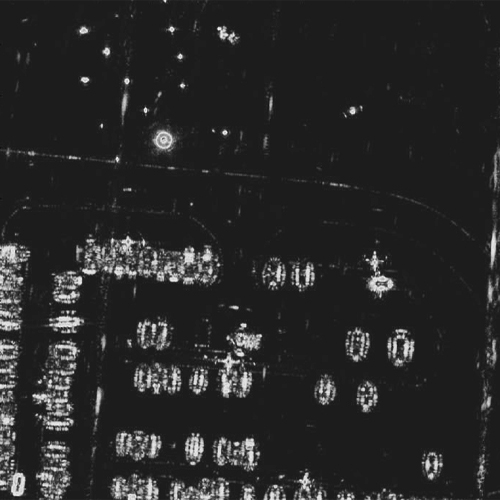}}
    \hfil
    \subfigure{\label{fig2b}\includegraphics[width=0.135\textwidth]{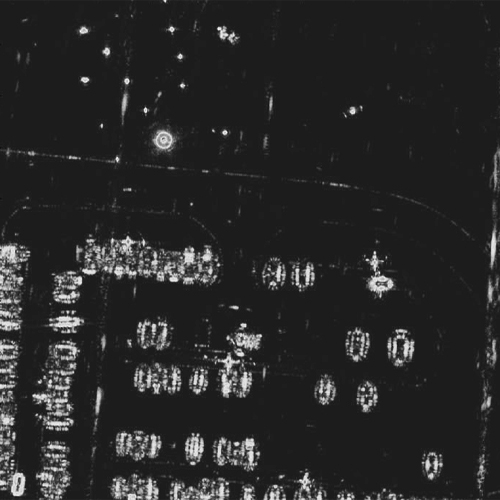}}
    \hfil
    \subfigure{\label{fig2b}\includegraphics[width=0.135\textwidth]{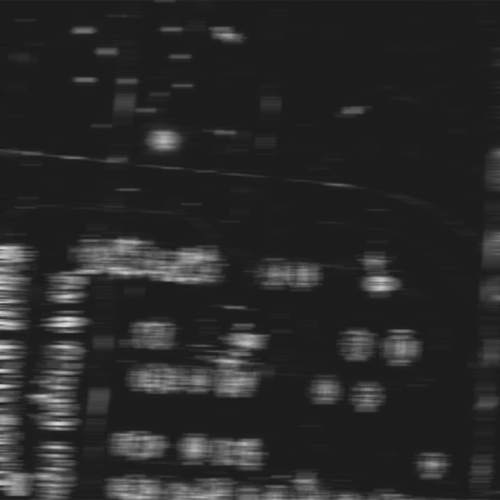}}
    \hfil
    \subfigure{\label{fig2b}\includegraphics[width=0.135\textwidth]{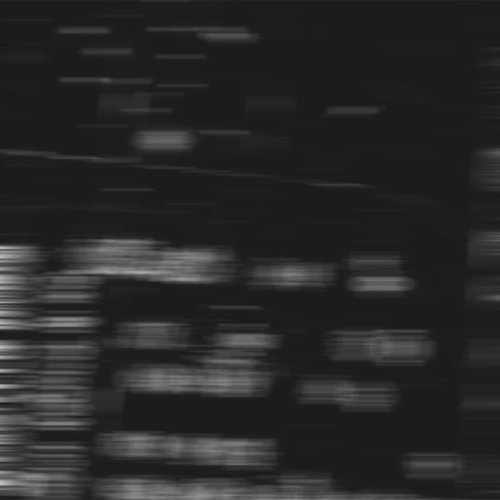}}
    \hfil
    \subfigure{\label{fig2b}\includegraphics[width=0.135\textwidth]{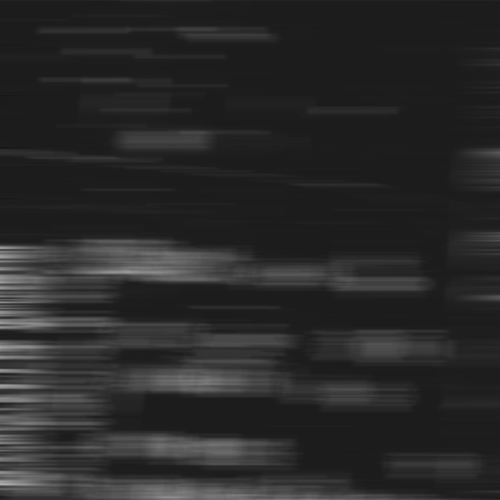}}

    \subfigure{\label{fig2b}\includegraphics[width=0.135\textwidth]{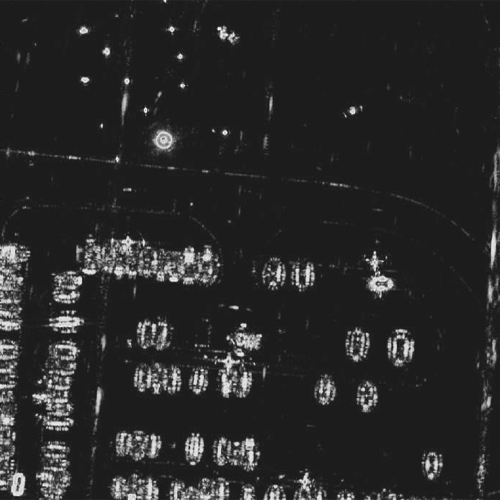}}
    \hfil
    \subfigure{\label{fig2b}\includegraphics[width=0.135\textwidth]{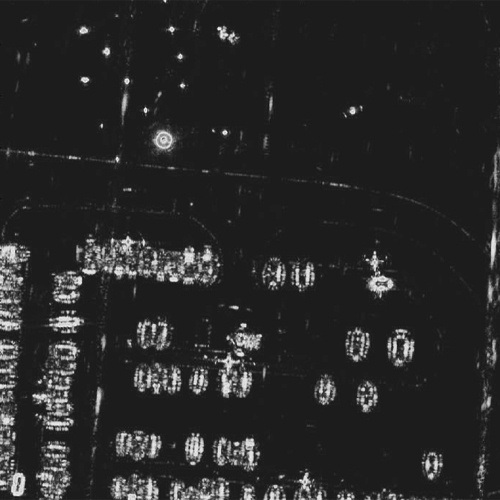}}
    \hfil
    \subfigure{\label{fig2b}\includegraphics[width=0.135\textwidth]{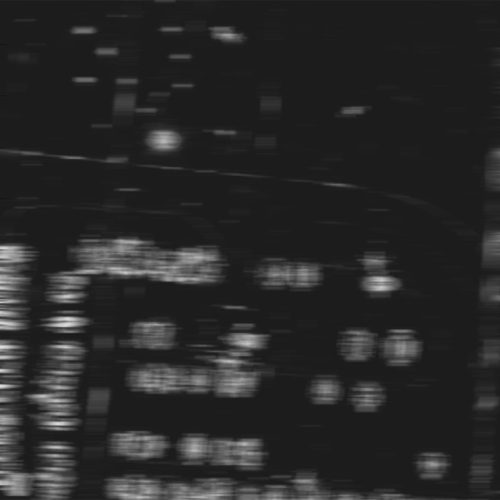}}
    \hfil
    \subfigure{\label{fig2b}\includegraphics[width=0.135\textwidth]{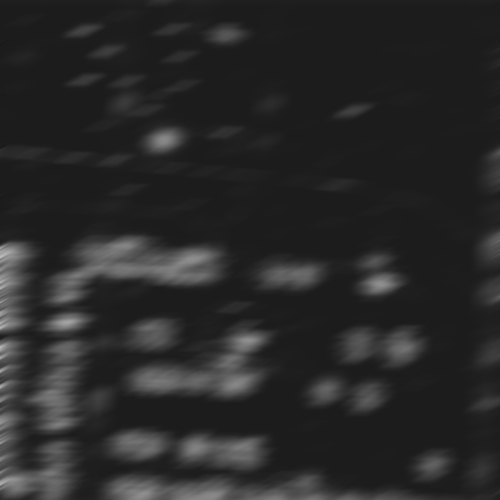}}
    \hfil
    \subfigure{\label{fig2b}\includegraphics[width=0.135\textwidth]{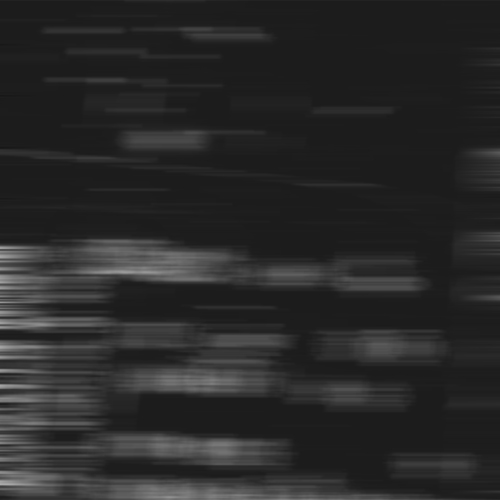}}
    \caption{Imaging results of different networks under training strategy 1. Columns 1 to 5 correspond to reconstructed results of $d = 0$ and $r' = 0$, $d = 0.18$ and $r’ = 1.05$, $d = 0.31$ and $r‘ = 2.04$, $d = 0.45$ and $r’ = 2.97$, $d = 0.54$ and $r‘ = 4.01$ for the testing data, respectively. Rows 1 to 5 correspond to REST4, LISTA4, LISTA7, ADMM-Net4 and ADMM-Net7, respectively.}\label{figr3}
\end{figure}

\begin{figure}[h]
    \centering
    \subfigure{\label{fig2b}\includegraphics[width=0.135\textwidth]{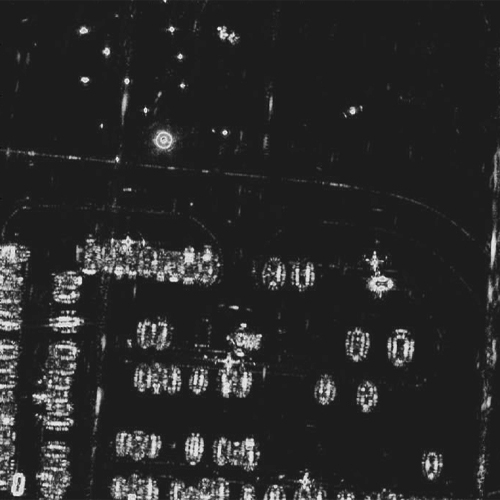}}
    \hfil
    \subfigure{\label{fig2b}\includegraphics[width=0.135\textwidth]{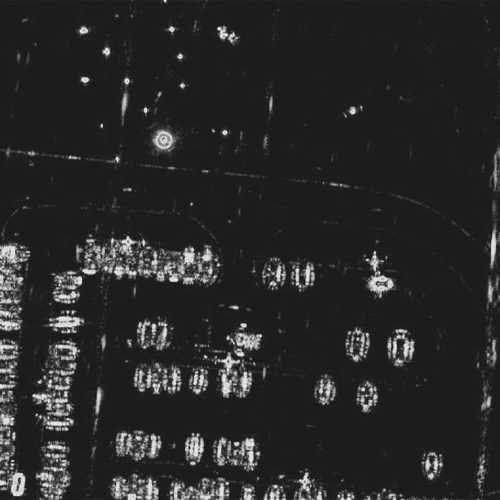}}
    \hfil
    \subfigure{\label{fig2b}\includegraphics[width=0.135\textwidth]{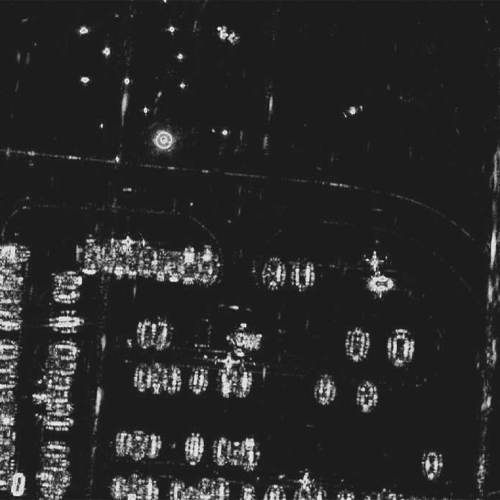}}
    \hfil
    \subfigure{\label{fig2b}\includegraphics[width=0.135\textwidth]{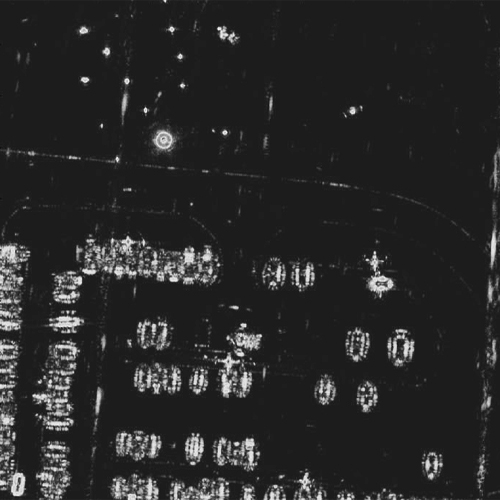}}
    \hfil
    \subfigure{\label{fig2b}\includegraphics[width=0.135\textwidth]{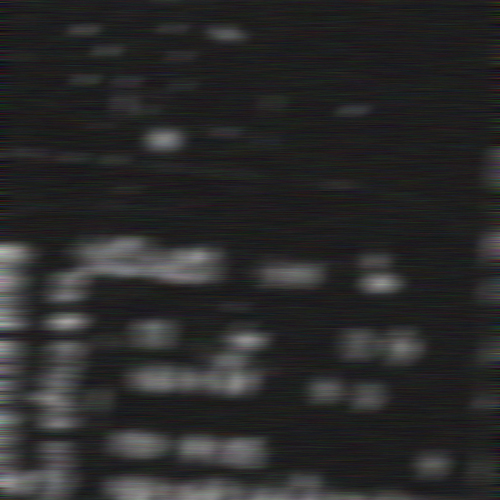}}

    \subfigure{\label{fig2b}\includegraphics[width=0.135\textwidth]{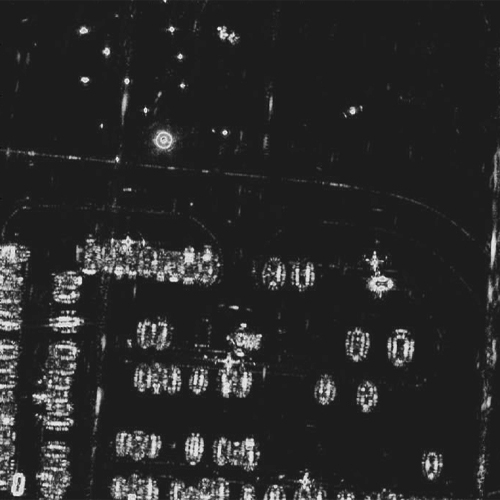}}
    \hfil
    \subfigure{\label{fig2b}\includegraphics[width=0.135\textwidth]{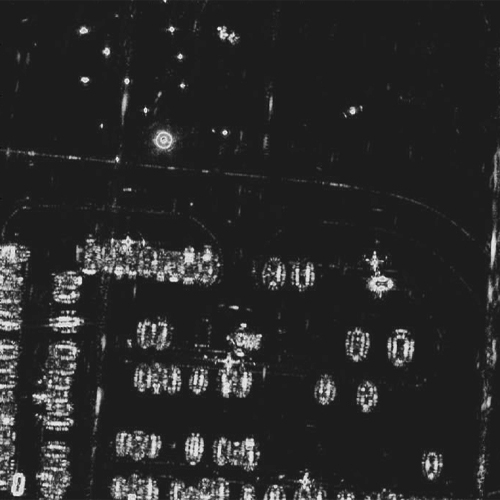}}
    \hfil
    \subfigure{\label{fig2b}\includegraphics[width=0.135\textwidth]{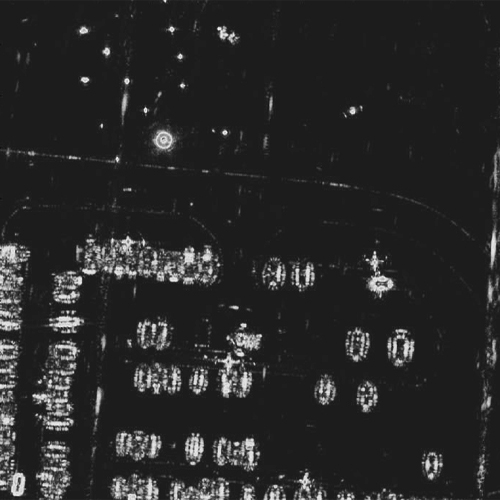}}
    \hfil
    \subfigure{\label{fig2b}\includegraphics[width=0.135\textwidth]{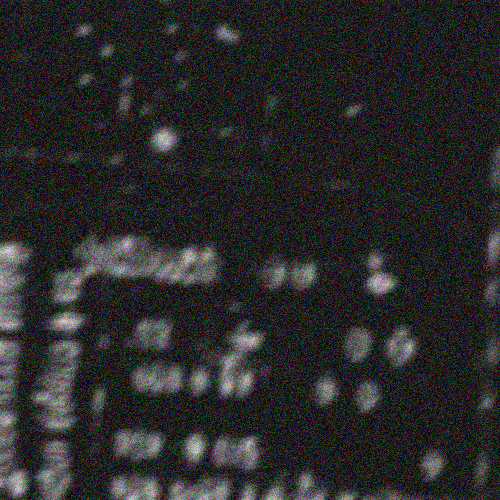}}
    \hfil
    \subfigure{\label{fig2b}\includegraphics[width=0.135\textwidth]{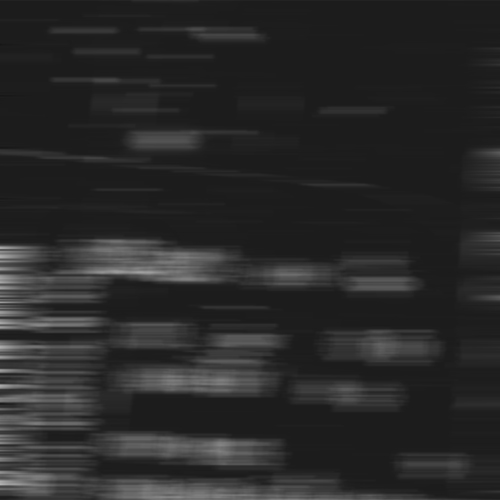}}

    \subfigure{\label{fig2b}\includegraphics[width=0.135\textwidth]{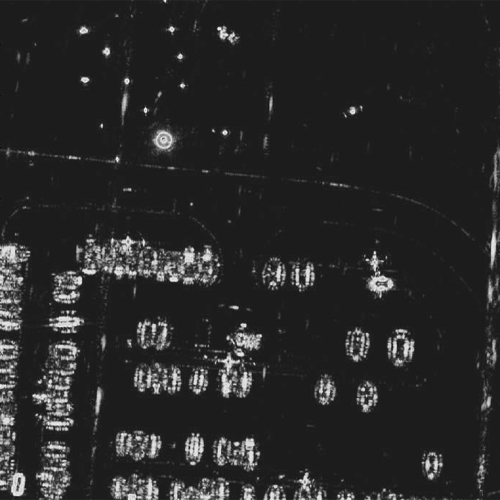}}
    \hfil
    \subfigure{\label{fig2b}\includegraphics[width=0.135\textwidth]{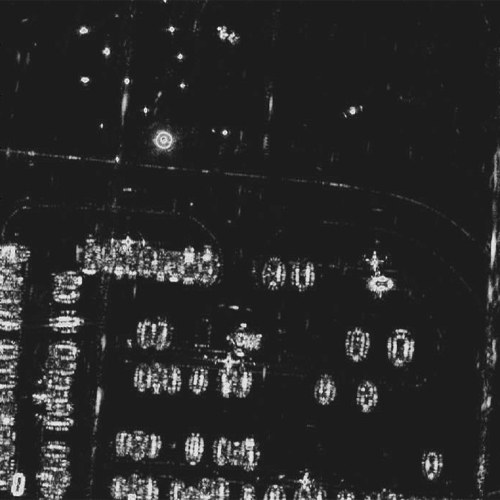}}
    \hfil
    \subfigure{\label{fig2b}\includegraphics[width=0.135\textwidth]{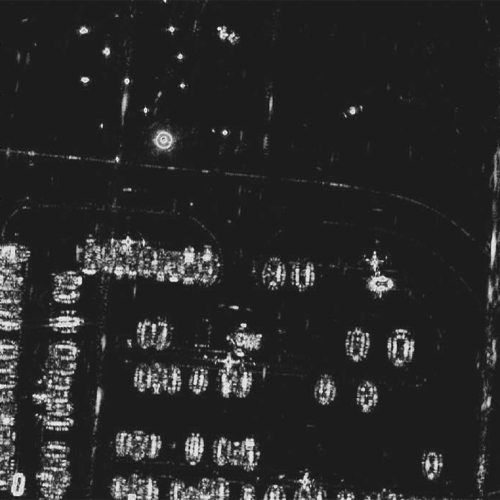}}
    \hfil
    \subfigure{\label{fig2b}\includegraphics[width=0.135\textwidth]{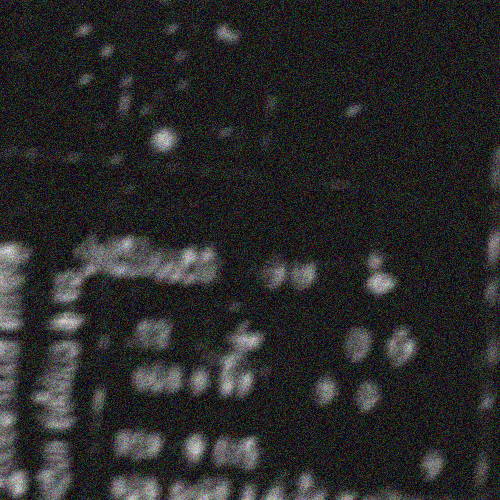}}
    \hfil
    \subfigure{\label{fig2b}\includegraphics[width=0.135\textwidth]{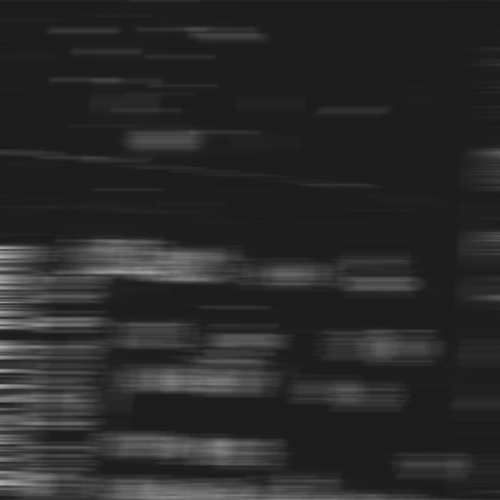}}

    \subfigure{\label{fig2b}\includegraphics[width=0.135\textwidth]{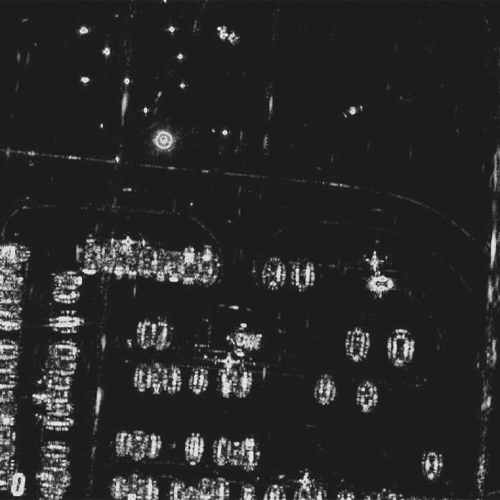}}
    \hfil
    \subfigure{\label{fig2b}\includegraphics[width=0.135\textwidth]{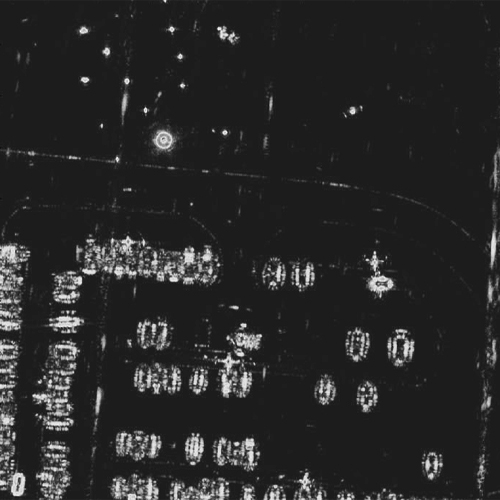}}
    \hfil
    \subfigure{\label{fig2b}\includegraphics[width=0.135\textwidth]{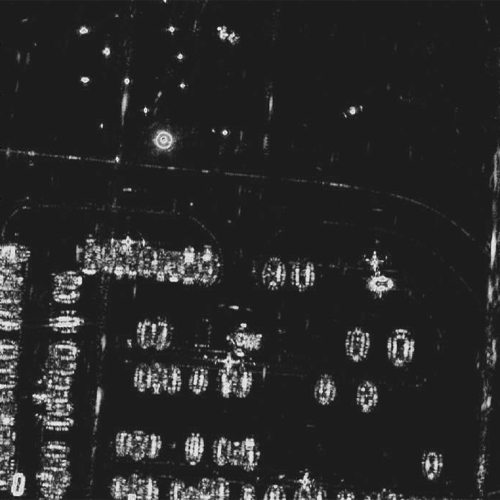}}
    \hfil
    \subfigure{\label{fig2b}\includegraphics[width=0.135\textwidth]{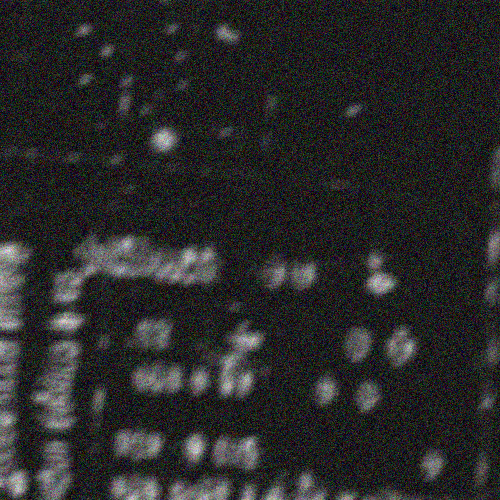}}
    \hfil
    \subfigure{\label{fig2b}\includegraphics[width=0.135\textwidth]{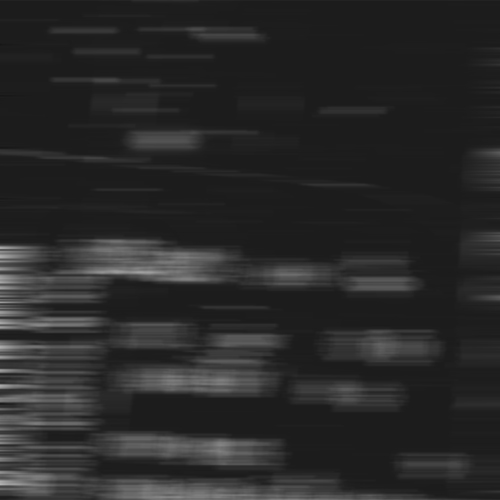}}

    \subfigure{\label{fig2b}\includegraphics[width=0.135\textwidth]{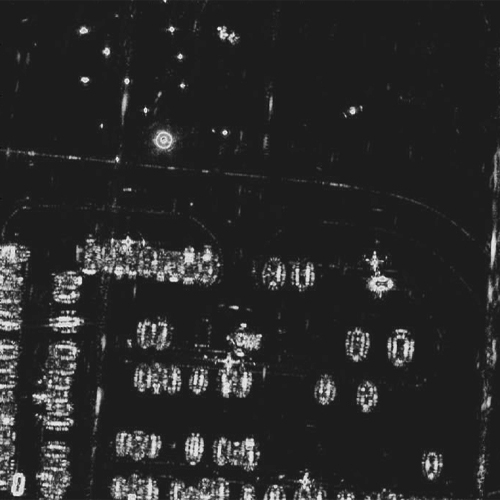}}
    \hfil
    \subfigure{\label{fig2b}\includegraphics[width=0.135\textwidth]{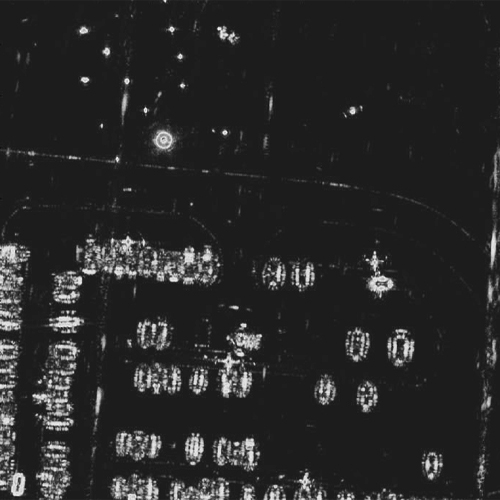}}
    \hfil
    \subfigure{\label{fig2b}\includegraphics[width=0.135\textwidth]{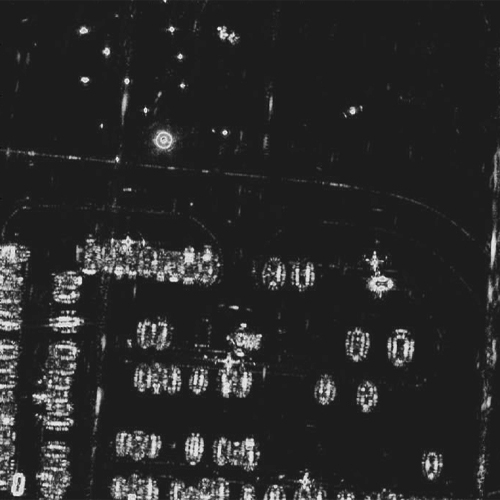}}
    \hfil
    \subfigure{\label{fig2b}\includegraphics[width=0.135\textwidth]{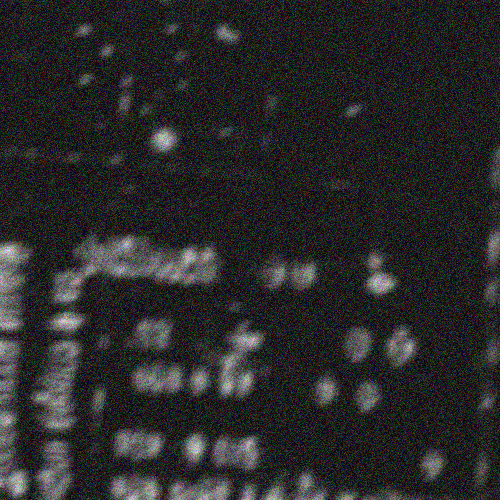}}
    \hfil
    \subfigure{\label{fig2b}\includegraphics[width=0.135\textwidth]{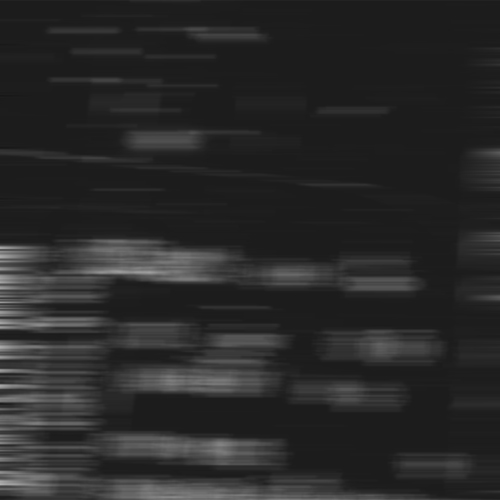}}
    \caption{Imaging results of different networks under training strategy 2. Columns 1 to 5 correspond to reconstructed results of $d = 0$ and $r' = 0$, $d = 0.18$ and $r’ = 1.05$, $d = 0.31$ and $r‘ = 2.04$, $d = 0.45$ and $r’ = 2.97$, $d = 0.54$ and $r‘ = 4.01$ for the testing data, respectively. Rows 1 to 5 correspond to REST4, LISTA7, LISTA7, ADMM-Net4 and ADMM-Net7, respectively.}\label{figr4}
\end{figure}

\begin{figure}[h]
    \centering
    \subfigure{\label{fig2b}\includegraphics[width=0.135\textwidth]{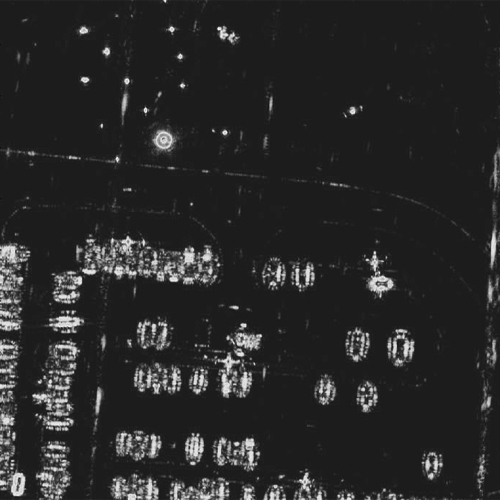}}
    \hfil
    \subfigure{\label{fig2b}\includegraphics[width=0.135\textwidth]{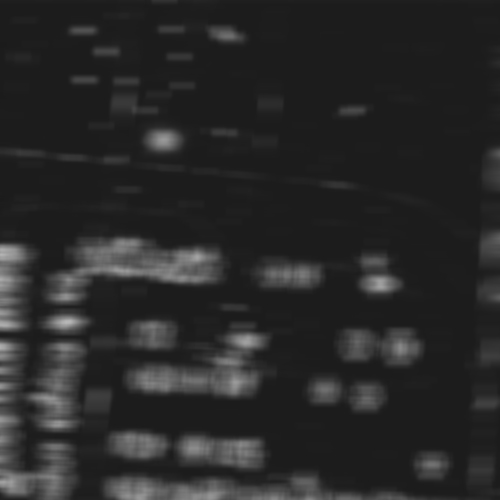}}
    \hfil
    \subfigure{\label{fig2b}\includegraphics[width=0.135\textwidth]{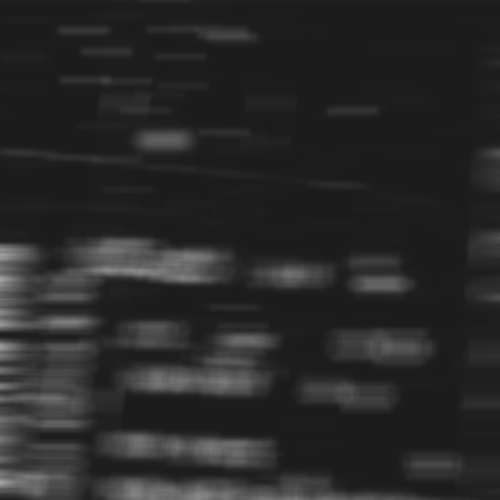}}
    \hfil
    \subfigure{\label{fig2b}\includegraphics[width=0.135\textwidth]{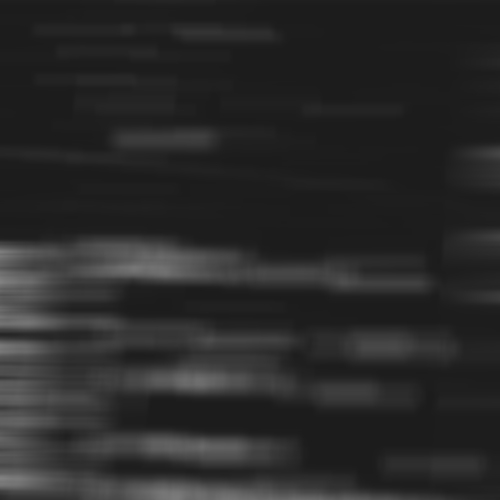}}
    \hfil
    \subfigure{\label{fig2b}\includegraphics[width=0.135\textwidth]{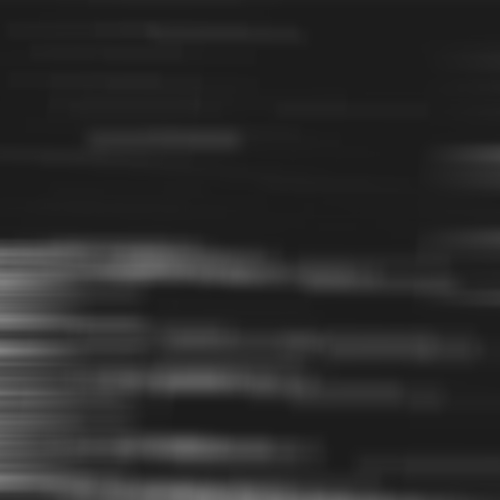}}

    \subfigure{\label{fig2b}\includegraphics[width=0.135\textwidth]{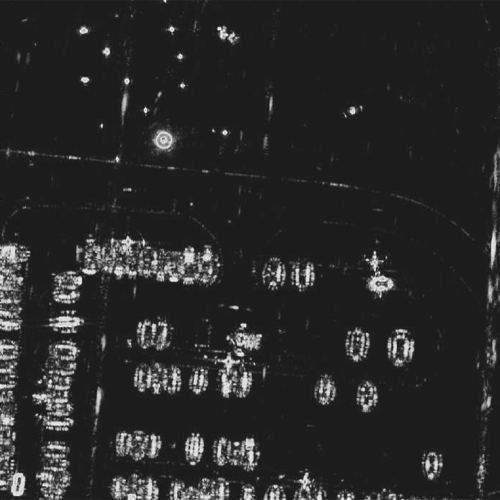}}
    \hfil
    \subfigure{\label{fig2b}\includegraphics[width=0.135\textwidth]{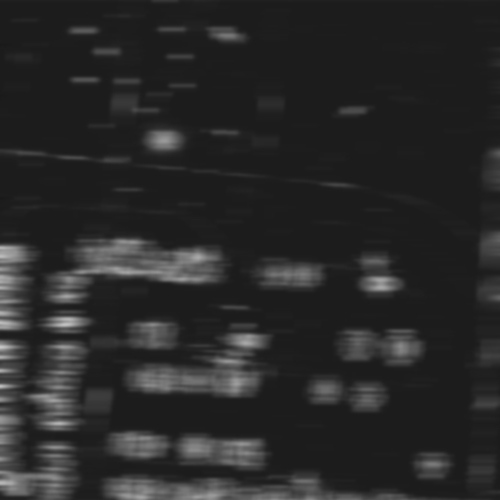}}
    \hfil
    \subfigure{\label{fig2b}\includegraphics[width=0.135\textwidth]{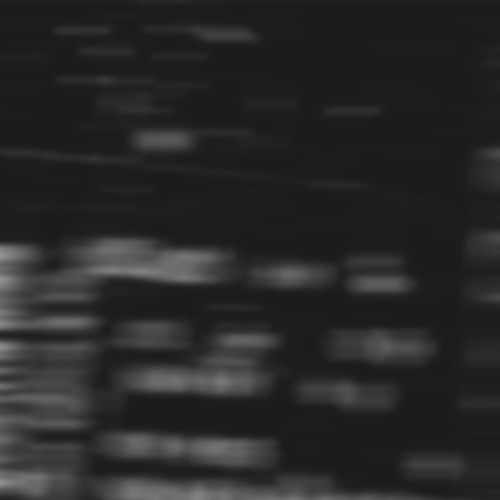}}
    \hfil
    \subfigure{\label{fig2b}\includegraphics[width=0.135\textwidth]{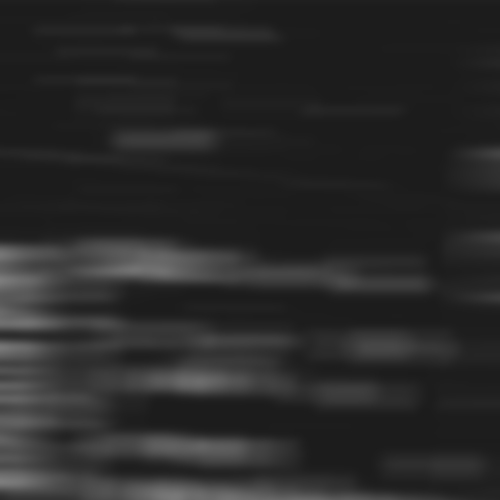}}
    \hfil
    \subfigure{\label{fig2b}\includegraphics[width=0.135\textwidth]{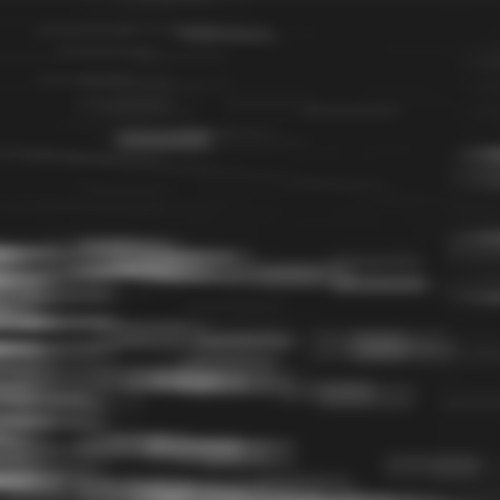}}

    \subfigure{\label{fig2b}\includegraphics[width=0.135\textwidth]{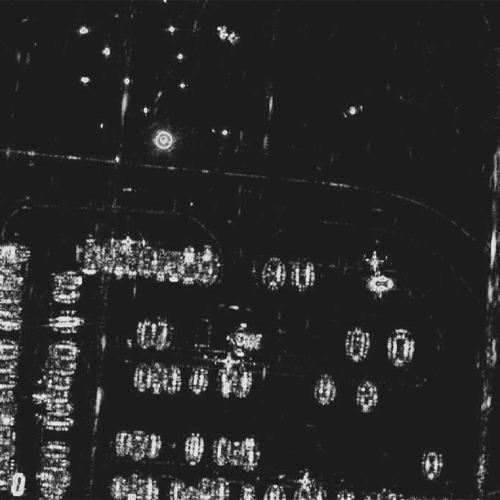}}
    \hfil
    \subfigure{\label{fig2b}\includegraphics[width=0.135\textwidth]{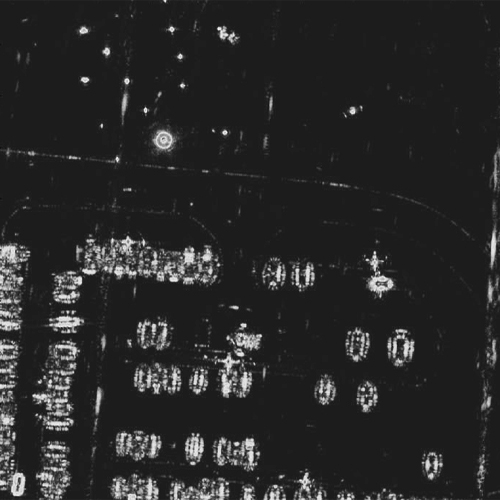}}
    \hfil
    \subfigure{\label{fig2b}\includegraphics[width=0.135\textwidth]{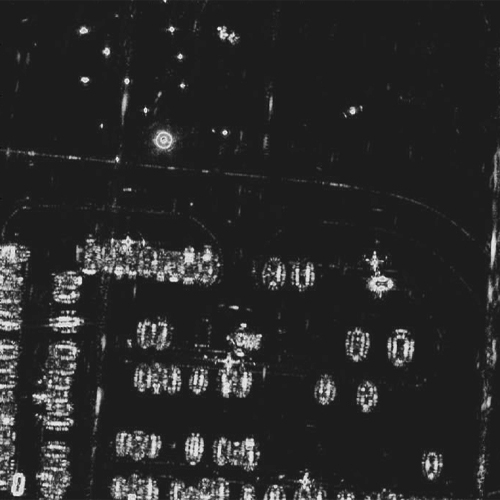}}
    \hfil
    \subfigure{\label{fig2b}\includegraphics[width=0.135\textwidth]{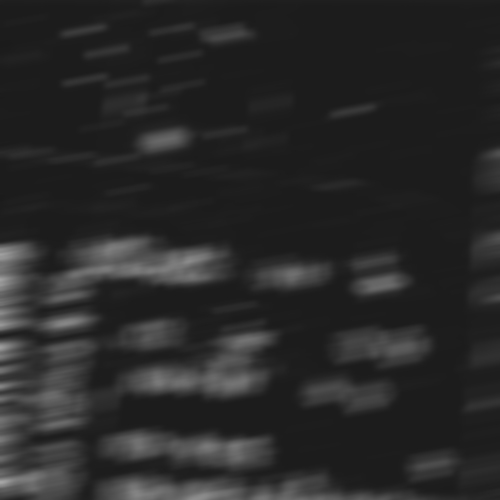}}
    \hfil
    \subfigure{\label{fig2b}\includegraphics[width=0.135\textwidth]{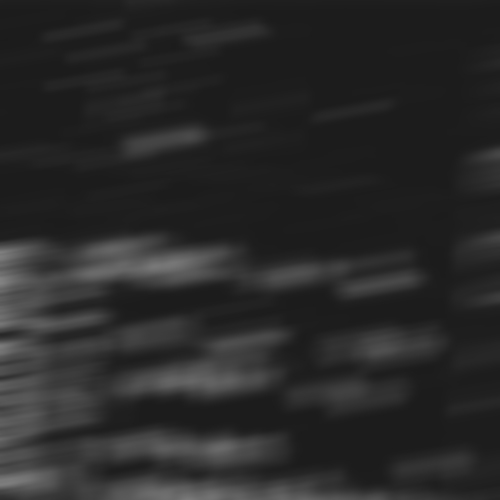}}

    \subfigure{\label{fig2b}\includegraphics[width=0.135\textwidth]{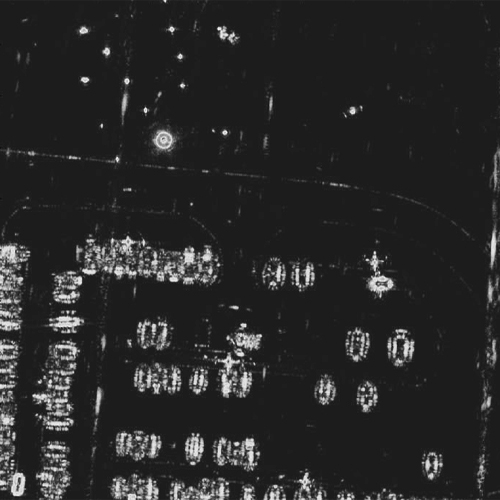}}
    \hfil
    \subfigure{\label{fig2b}\includegraphics[width=0.135\textwidth]{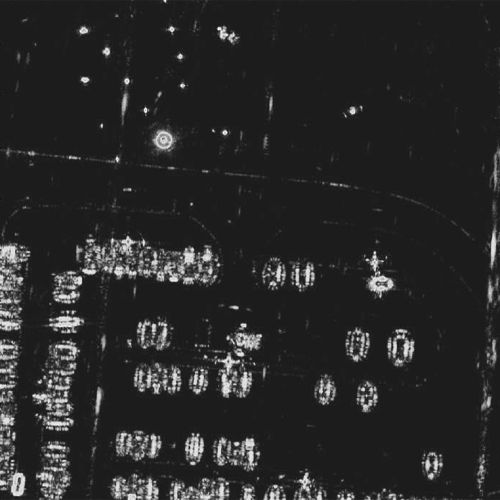}}
    \hfil
    \subfigure{\label{fig2b}\includegraphics[width=0.135\textwidth]{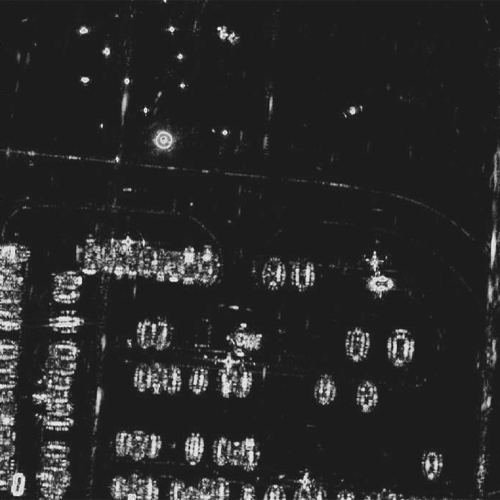}}
    \hfil
    \subfigure{\label{fig2b}\includegraphics[width=0.135\textwidth]{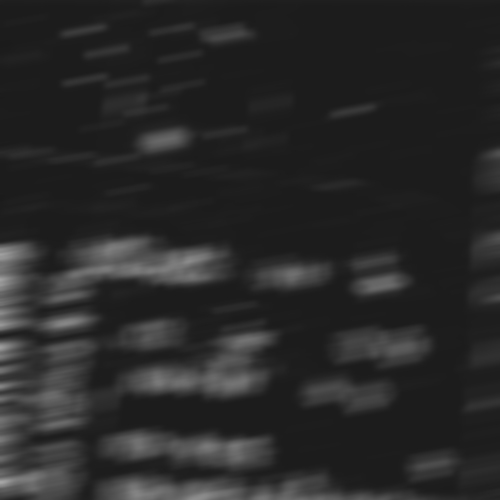}}
    \hfil
    \subfigure{\label{fig2b}\includegraphics[width=0.135\textwidth]{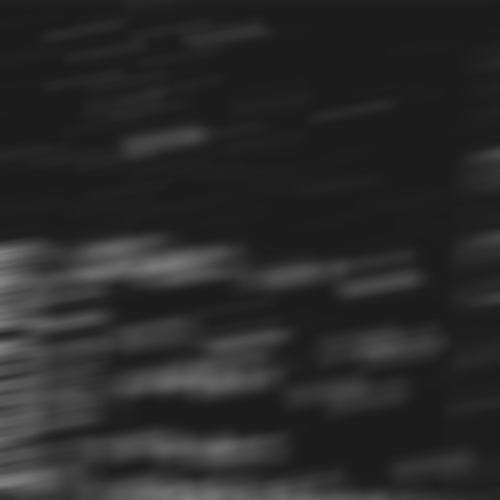}}
    \caption{Imaging results of different model-based approaches. Columns 1 to 5 correspond to reconstructed results of $d = 0$ and $r' = 0$, $d = 0.18$ and $r’ = 1.05$, $d = 0.31$ and $r‘ = 2.04$, $d = 0.45$ and $r’ = 2.97$, $d = 0.54$ and $r‘ = 4.01$ for the testing data, respectively. Rows 1 to 5 correspond to BP, ISTA, robust ISTA and MU-GAMP algorithms, respectively.}\label{figr5}
\end{figure}

The imaging results of different networks under training strategy 1 and training strategy 2 are shown in Fig. \ref{figr3} and \ref{figr4}, respectively.
Note that in Fig. \ref{figr3} and \ref{figr4}, in line with standard practice, we present the amplitude value of $\sigma$ and we combined the small images of testing data into a whole large image in order to better display the results.
Imaging results by model-based approaches, i.e., BP, ISTA, robust ISTA and MU-GAMP, are shown in Fig. \ref{figr5}. In Fig. \ref{figr3}, \ref{figr4} and \ref{figr5}, columns 1 to 5 correspond to reconstructed results of $d = 0$ and $r'= 0$, $d = 0.18$ and $r'= 1.05$, $d = 0.31$ and $r'= 2.04$, $d = 0.45$ and $r'= 2.97$, $d = 0.54$ and $r'= 4.01$ for the testing data, respectively. In Fig. \ref{figr3} and \ref{figr4}, Rows 1 to 5 correspond to REST4, LISTA4, LISTA7, ADMM-Net4 and ADMM-Net7, respectively. In Fig. \ref{figr5}, rows 1 to 4 correspond to BP, ISTA, robust ISTA and MU-GAMP algorithms, respectively.

From Fig. \ref{figr3}, \ref{figr4} and \ref{figr5}, we observe that for the learning-based approaches like REST, LISTA and ADMM-Net, the performance is better when the networks are trained in strategy 2. It shows that when mismatch is taken into consideration during training, the robustness of the networks to mismatch error is improved.
In strategy 1, the reconstructed results of LISTA and ADMM-Net become seriously blurred when $r' = 2.04$, while the results of REST are blurred when $r' = 2.97$. This is due to the fact that the REST is especially designed for inverse problem with mismatch even when there is no attempt to robustify the network during training.
In strategy 2, the reconstructed results of LISTA and ADMM-Net are seriously blurred when $r' = 2.97$, while results of REST are blurred when $r' = 4.01$. When $r'= 2.97$ and $r'= 4.01$, results of REST are much better than those of LISTA and ADMM-Net under the same training strategy. This shows that REST is also much more immune to model mismatch even when one attempts to robustify the various networks during training.

Robust ISTA and MU-GAMP have better performance than BP and ISTA. However, the proposed REST outperforms Robust ISTA and MU-GAMP especially in the cases $r' = 2.97$ and  $r' = 4.01$.

Next, the image qualities corresponding to different algorithms and networks are measured quantitatively, using MSE of target vector $x$ to evaluate the reconstruction performance.
\renewcommand{\arraystretch}{1.2}
\begin{table}[tp]
  \centering
  \fontsize{6.5}{6}\selectfont
  \begin{threeparttable}
  \caption{MSE of different approaches with different $r'$. ($r=0$ for strategy T1 and $r = 3.01$ for strategy T2)}\label{T4}
    \begin{tabular}{cccccc}
    \toprule
    \multirow{2}{*}{Methods}&
    \multicolumn{5}{c}{MSE}\cr
    \cmidrule(lr){2-6}
    &$r' = 0$&$r' = 1.05$&$r' = 2.04$&$r' = 2.97$&$r' = 4.01$\cr
    \midrule
    REST4(T1)&0.0054&0.0077&0.0096&0.046&0.10\cr
    REST4(T2)&0.0077&0.0084&0.0097&0.013&0.074\cr
    LISTA4(T1)&0.0011&0.012&0.096&0.17&0.29\cr
    LISTA4(T2)&0.0069&0.0086&0.0097&0.059&0.19\cr
    LISTA7(T1)&0.0010&0.017&0.087&0.16&0.24\cr
    LISTA7(T2)&0.0052&0.0081&0.0094&0.051&0.16\cr
    ADMM-Net4(T1)&0.00097&0.0096&0.088&0.16&0.24\cr
    ADMM-Net4(T2)&0.0074&0.082&0.0096&0.056&0.18\cr
    ADMM-Net7(T1)&0.00095&0.0097&0.087&0.15&0.22\cr
    ADMM-Net7(T2)&0.0072&0.0080&0.0093&0.050&0.16\cr
    BP&0.0025&0.058&0.092&0.29&0.48\cr
    ISTA&0.0022&0.063&0.095&0.25&0.44\cr
    Robust ISTA&0.0069&0.0097&0.032&0.097&0.21\cr
    MU-GAMP&0.0077&0.0092&0.029&0.090&0.19\cr
    \bottomrule
    \end{tabular}
    \end{threeparttable}
\end{table}
The performance of different approaches is listed in Table \ref{T4}, where T1 and T2 in the brackets represent that the networks are trained using strategy 1 and strategy 2, respectively. The observations obtained from Fig. \ref{figr3}, \ref{figr4} and \ref{figr5} could be verified by Table \ref{T4} as well. In addition, we also note that:
\begin{itemize}
\item LISTA7 and ADMM-Net7 have slightly better performance than LISTA4 and ADMM-Net4 especially when the mismatch in testing data is large. This indicates that increasing the number of layers in LISTA and ADMM-Net networks can improve the robustness, however, the improvement is very limited.
\item The proposed REST network is much more robust to the mismatch error than the existing learning- and model-based approaches. Even REST trained in training strategy 1 has a better performance than LISTA and ADMM-Net trained in training strategy 2 and other model-based approaches.
\end{itemize}


\section{Conclusion}

Deep learning has achieved significant successes in various signal and image processing tasks, but it is also vulnerable to various perturbations. By adopting algorithm unrolling techniques and proposing a REST network, we show that it is possible to design neural network architectures that can reliably recover high-dimensional sparse vectors from low-dimensional noisy linear measurements in the presence of challenging sample-wise model mismatch. We also apply the proposed REST network to solve the synthetic aperture radar imaging problem, wherein model mismatch occurs due to motion uncertainty of the moving radar platform. Experiments on both compressive sensing and synthetic aperture radar imaging tasks show the proposed REST network outperforms state-of-the-art learning-based and model-based approaches in view of various additional operations that seem to endow the architecture with inherent robustness to mismatches.

\section{Appendix A}
This appendix briefly introduces the performance of the proposed REST network with shared or unshared weights in each layer. We run various experiments (in line with the experimental set-up reported in Section IV-A).
In Table \ref{TA}, we compare the testing loss ($10\log_{10}$MSE) of shared and unshared parameters of REST with different $r$ and $r'$ with 1000 training samples, 100 testing samples and 2000 epochs, where REST-I and REST-II denote the shared and unshared parameters of REST, respectively. It can be observed from Table \ref{TA} that the networks with shared and unshared parameters have very similar performance. When the mismatch in testing data is large, i.e., $r' = 8$ and $r' = 12$, the shared parameters network has slightly better performance. 

\begin{table*}[t]
  \centering
  \fontsize{6.5}{6}\selectfont
  \begin{threeparttable}
  \caption{Testing loss ($10\log_{10}$MSE) of shared and unshared parameters of REST with different $r$ and $r'$ with 1000 training samples, 100 testing samples and 2000 epochs.}\label{TA}
    \begin{tabular}{ccccccccccccc}
    \toprule
    \multirow{3}{*}{Methods}&
    \multicolumn{4}{c}{ $r = 0$}&\multicolumn{4}{c}{$r = 4$}&\multicolumn{4}{c}{$r= 8$}\cr
    \cmidrule(lr){2-5} \cmidrule(lr){6-9} \cmidrule(lr){10-13}
    &$r' = 0$&$r' = 4$&$r' = 8$&$r' = 12$&$r' = 0$&$r' = 4$&$r' = 8$&$r' = 12$&$r' = 0$&$r' = 4$&$r' = 8$&$r' = 12$\cr
    \midrule
    REST-I&-2.57&-1.58&-0.99&-0.74&-2.40&-1.72&-1.22&-1.01&-2.05&-1.87&-1.52&-1.25\cr
    REST-II&-2.62&-1.60&-0.99&-0.78&-2.41&-1.70&-1.23&-1.01&-2.01&-1.86&-1.51&-1.23\cr
    \bottomrule
    \end{tabular}
    \end{threeparttable}
\end{table*}

\section{Appendix B}
This appendix briefly overviews other options to turn the robust ISTA algorithm appearing in (\ref{e10}) onto other neural network structures, depending on how one selects the learnable parameters.

{\bf Option 1:} This option is based on re-writing (\ref{e10}) as follows:
        \begin{align}
        \label{e13}
        x^{k+1} = {\cal S}_{\mu \lambda}
        ( A_1 x^{k}  +  A_2 y),
        \end{align}
        where $A_1 = I + \frac{2\mu\| y - Ax^k \|_2^2I} {(1+ \|{  x^k} \|_2^2)^2} - \frac{2\mu A^TA} {1+ \|{  x^k} \|_2^2 }$ and $A_2 = \frac{2\mu A^T} {1+ \|{  x^k} \|_2^2  }$. We then set $\mu \lambda$, $A_1$ and $A_2$ as learnable parameters. Clearly, this leads to an unfolded network akin to LISTA, so this unfolding approach does not carry any advantage.

{\bf Option 2:} This option is based on re-writing (\ref{e10}) as follows:
        \begin{align}
        \label{e14}
        x^{k+1} = {\cal S}_{\mu \lambda} ( x^{k} &+ \frac{2\mu_1} {(1+ \|{  x^k} \|_2^2)^2} x^{k} \nonumber\\
         &- \frac{2 \mu A_1} {1+ \|{  x^k} \|_2^2 } x^{k}  +  \frac{2 \mu A_2} {1+ \|{  x^k} \|_2^2  } y )
        \end{align}
        where $\mu_1 = \mu \| y - Ax^k \|_2^2$ and we have replaced $ A^TA$ in the third term by $A_1$, and $A^T$ in the fourth term by $A_2$. Here, we set $\mu \lambda$, $\mu_1$, $A_1$ and $A_2$ as learnable parameters.

{\bf Option 3:} This option is based on re-writing (\ref{e10}) as follows:
        \begin{align}
        \label{e15}
        x^{k+1} = {\cal S}_{\mu \lambda} ( x^{k} & + \frac{2\mu\| y - A_1x^k \|_2^2} {(1+ \|{  x^k} \|_2^2)^2} x^{k} \nonumber\\
        & - \frac{2 \mu A_2} {1+ \|{  x^k} \|_2^2 } x^{k}  +  \frac{2 \mu A_3} {1+ \|{  x^k} \|_2^2  } y )
        \end{align}
        where we have replaced $A$ in the second term by $A_1$, $ A^TA$ in the third term by $A_2$, and $ A^T$ in the fourth term by $A_3$. Here, we set $\lambda$, $\mu$, $A_1$, $A_2$ and $A_3$ as learnable parameters. The option 3 leads to a network structure akin to the one deriving from (\ref{e10}) and (\ref{e11}).

{\bf Option 4:} Finally, one can also re-write (\ref{e10}) as follows:
        \begin{align}
        \label{e16}
        x^{k+1} = {\cal S}_{\mu \lambda} ( x^{k} &+ \frac{2\mu y^Tyx^{k} } {(1+ \|{  x^k} \|_2^2)^2 } \nonumber\\
        &-\frac{ 4\mu y^TA_1x^kx^{k} } {(1+ \|{  x^k} \|_2^2)^2 } +\frac{  2\mu x^{kT}A_2x^{k}x^{k}  } {(1+ \|{  x^k} \|_2^2)^2 }\nonumber\\
         &-\frac{   2\mu A_2 x^k  } {1+ \|{  x^k} \|_2^2 }+ \frac{   2\mu  A_3 y    } {1+ \|{  x^k} \|_2^2 })
        \end{align}
        where we have replaced the matrix $ A$ in the third term by $A_1$, $A^TA$ in the fourth and fifth terms by $A_2$, and $ A^T$ in the sixth term by $A_3$. Here, we set $\lambda$, $\mu$, $A_1$, $A_2$ and $A_3$ as learnable parameters.

We run various experiments (in line with the experimental set-up reported in Section IV-A) in order to test the performance of the different unfolding options.
In Table \ref{TB2}, we compare the testing loss for different unfolding strategies of REST, where REST2-6, REST2-8, REST3-6 and REST4-6 represent the different unfolding strategies shown in subsection III-B. The definitions are shown in Table \ref{TB1}. Note that REST2-8 has similar parameter number as those of REST3-6 and REST4-6, i.e., 7616 parameters for REST2-8 and 7662 parameters for REST3-6 and REST4-6. It could be observed from Table \ref{TB2} that the networks unfolded by the third and fourth schemes have similar performance, while the performance of the network by the second unfolding strategy is slightly worse. This motivated us to adopt the unfolding approach deriving from (\ref{e10}) and (\ref{e11}).
\renewcommand{\arraystretch}{1.0}
\begin{table}[tp]
  \centering
  \fontsize{6.5}{6}\selectfont
  \begin{threeparttable}
  \caption{Different unfolding strategies for REST.}\label{TB1}
    \begin{tabular}{ll}
    \toprule
    REST2-6&REST unfolded by the 2nd strategy with 6 layers\cr
    REST2-8&REST unfolded by the 2nd strategy with 8 layers\cr
    REST3-6&REST unfolded by the 3rd strategy with 6 layers\cr
    REST4-6&REST unfolded by the 4th strategy with 6 layers\cr
    \bottomrule
    \end{tabular}
    \end{threeparttable}
\end{table}
\renewcommand{\arraystretch}{1.2}
\begin{table*}[t]
  \centering
  \fontsize{6.5}{6}\selectfont
  \begin{threeparttable}
  \caption{Testing loss ($10\log_{10}$MSE) of different unfolding strategies of REST with different $r$ and $r'$ with 1000 training samples, 100 testing samples and 2000 epochs.}\label{TB2}
    \begin{tabular}{ccccccccccccc}
    \toprule
    \multirow{3}{*}{Methods}&
    \multicolumn{4}{c}{ $r = 0$}&\multicolumn{4}{c}{$r = 4$}&\multicolumn{4}{c}{$r= 8$}\cr
    \cmidrule(lr){2-5} \cmidrule(lr){6-9} \cmidrule(lr){10-13}
    &$r' = 0$&$r' = 4$&$r' = 8$&$r' = 12$&$r' = 0$&$r' = 4$&$r' = 8$&$r' = 12$&$r' = 0$&$r' = 4$&$r' = 8$&$r' = 12$\cr
    \midrule
    REST2-6&-2.92&-1.55&-0.90&-0.67&-2.76&-1.70&-1.11&-0.94&-2.69&-1.84&-1.23&-1.01\cr
    REST2-8&-2.94&-1.57&-0.94&-0.69&-2.76&-1.71&-1.14&-0.97&-2.73&-1.86&-1.24&-1.06\cr
    REST3-6&-2.57&-1.58&-0.99&-0.74&-2.40&-1.72&-1.22&-1.01&-2.05&-1.87&-1.52&-1.25\cr
    REST4-6&-2.59&-1.57&-0.99&-0.73&-2.41&-1.72&-1.22&-1.02&-2.05&-1.86&-1.52&-1.26\cr
    \bottomrule
    \end{tabular}
    \end{threeparttable}
\end{table*}

\section{Appendix C}
In this appendix, we run various experiments (in line with the experimental set-up reported in Section IV-A) to support our intuitions on the comparison between REST and LISTA. As mentioned in Section III-C, the main difference between a layer in LISTA and in REST lies in two additional processing operations: 1) one corresponds to the second term in (\ref{e11}) and 2) the other corresponds to the normalization operation by $1 + \|x\|^2_2$.

Here, we design two networks to be used for comparison with LISTA and REST.
The first one is to delete the second term in (\ref{e11}) leading to
\begin{align}
\label{eC1}
x^{k+1} =
{\cal S}_{\mu \lambda} \left (  x^{k}  - \frac{2\mu A_2} {1+ \|{  x^k} \|_2^2 } x^{k}  +  \frac{2\mu A_3} {1+ \|{  x^k} \|_2^2  } y\right) .
\end{align}
Compared with LISTA, this network has the same parameter number with an additional normalization by $1 + \|x\|^2_2$.

The second network reduces the normalization in (\ref{e11}) resulting in
\begin{align}
\label{eC2}
x^{k+1} =
{\cal S}_{\mu \lambda} \left (  x^{k} + {2\mu\| y - A_1x^k \|_2^2}  x^{k} - {2\mu A_2}  x^{k}  +  {2\mu A_3}  y\right) .
\end{align}
Compared with LISTA, the main difference is the second term in (\ref{eC2}), which enlarges the number of parameters.

\begin{table*}[t]
  \centering
  \fontsize{6.5}{6}\selectfont
  \begin{threeparttable}
  \caption{Testing loss ($10\log_{10}$MSE) of different networks with different $r$ and $r'$ with 1000 training samples, 100 testing samples and 2000 epochs.}\label{TC}
    \begin{tabular}{ccccccccccccc}
    \toprule
    \multirow{3}{*}{Methods}&
    \multicolumn{4}{c}{ $r = 0$}&\multicolumn{4}{c}{$r = 4$}&\multicolumn{4}{c}{$r= 8$}\cr
    \cmidrule(lr){2-5} \cmidrule(lr){6-9} \cmidrule(lr){10-13}
    &$r' = 0$&$r' = 4$&$r' = 8$&$r' = 12$&$r' = 0$&$r' = 4$&$r' = 8$&$r' = 12$&$r' = 0$&$r' = 4$&$r' = 8$&$r' = 12$\cr
    \midrule
    REST6 &-2.57&-1.58&-0.99&-0.74&-2.40&-1.72&-1.22&-1.01&-2.05&-1.87&-1.52&-1.25\cr
    NI6   &-2.77&-1.62&-0.69&-0.13&-2.44&-1.70&-0.87&-0.31&-1.88&-1.44&-1.92&-0.61\cr
    NI8   &-2.78&-1.61&-0.70&-0.13&-2.43&-1.71&-0.87&-0.32&-1.71&-1.45&-1.90&-0.60\cr
    NII6  &-3.50&-1.77&-0.73&-0.22&-2.78&-1.32&-1.08&-0.47&-1.97&-1.58&-0.97&-0.66\cr
    LISTA6&-3.49&-0.62&0.48&0.54&-2.57&-0.74&-0.26&0.09&-1.60&-0.78&-0.47&-0.22\cr
    LISTA8&-3.49&-0.65&0.47&0.52&-2.56&-0.79&-0.27&0.07&-1.62&-0.81&-0.53&-0.27\cr
    \bottomrule
    \end{tabular}
    \end{threeparttable}
\end{table*}
In Table \ref{TC}, we compare the testing loss ($10\log_{10}$MSE) of different networks mentioned above with different $r$ and $r'$ with 1000 training samples, 100 testing samples and 2000 epochs, where REST6, LISTA 6 and LISTA8 are defined in Table \ref{T0} and NI6, NI8 and NII8 are the network in (\ref{eC1}) with 6 layers, network in (\ref{eC1}) with 8 layers and the network in (\ref{eC2}) with 6 layers, respectively.
REST6, NI8 and LISTA8 have similar parameter numbers, i.e., 765,012 parameters for REST6, 760,008 parameters for LISTA8 and NI8. However, the numbers of learnable parameters are different, i.e., 127,502 learnable parameters for REST6, 95,001 learnable parameters for LISTA8 and NI8.
It could be observed from Table \ref{TC} that both NI and NII have smaller testing errors than LISTA and larger testing errors than REST when there is model mismatch in testing data, which indicates that both two additional processing operations promote the robustness of the network.

\end{document}